# Design description of

# Wisdom Computing Persperctive

Author: TianYi Yu

Name: rainbow_yu

GitHub: https://github.com/rainbowyuyu

April 2025

# catalogue

# abstract

This course design aims to develop and research a handwriting matrix recognition and step-by-step visual calculation process display system, addressing the issue of abstract formulas and complex calculation steps that students find difficult to understand when learning mathematics. By integrating artificial intelligence with visualization animation technology, the system enhances precise recognition of handwritten matrix content through the introduction of Mamba backbone networks, completes digital extraction and matrix reconstruction using the YOLO model, and simultaneously combines CoordAttention coordinate attention mechanisms to improve the accurate grasp of character spatial positions. The calculation process is demonstrated frame by frame through the Manim animation engine, vividly showcasing each mathematical calculation step, helping students intuitively understand the intrinsic logic of mathematical operations. Through dynamically generating animation processes for different computational tasks, the system exhibits high modularity and flexibility, capable of generating various mathematical operation examples in real-time according to student needs. By innovating human-computer interaction methods, it brings mathematical calculation processes to life, helping students bridge the gap between knowledge and understanding on a deeper level, ultimately achieving a learning experience where "every step is understood." The system's scalability and interactivity make it an intuitive, user-friendly, and efficient auxiliary tool in education.

# 1 foreword

## 1.1motive

In today's higher education system, which increasingly emphasizes competency and practical application, students not only need to accumulate knowledge but also face the challenge of understanding and applying this knowledge. This is especially true in mathematics courses, where many students are often bogged down by complex formulas and computational steps, unable to truly grasp the underlying reasoning. Particularly in linear algebra and matrix computation courses, students lack an intuitive understanding of formulas and frequently make mistakes during calculations, failing to effectively build a systematic knowledge framework. This issue is particularly prominent in the current educational environment, especially in advanced courses, where students' understanding remains superficial, lacking deep reflection and practice. With the development of information technology, there are tools available on the market such as Microsoft Math and MathDF that provide some computational support, but these tools mostly focus on providing final results, neglecting students 'need for a detailed understanding of the calculation process. More importantly, these tools lack personalization and interactivity, failing to offer customized support based on different students' learning progress and comprehension levels.

## 1.2The problem to be solved

Data Set Generation In the field of mathematics learning and computation, especially in the teaching of matrix operations, existing datasets mostly focus on areas such as image recognition and text classification, with relatively few dedicated to mathematical matrix recognition. This presents certain challenges when constructing training sets. Particularly for handwritten matrix recognition, the lack of datasets results in insufficient coverage of different writing styles and matrix sizes.

The accuracy of image recognition: Existing object detection algorithms, such as YOLO (You Only Look Once) and [1], have been widely applied to tasks like object detection and image classification, demonstrating excellent performance in real-time processing and accuracy. However, image recognition of handwritten matrices faces numerous challenges, including font variations, noise interference, and image blurriness, all of which can affect the precision of recognition.

Mathematical Expression Localization After completing the recognition of matrix images, the next major challenge is to accurately locate the identified elements in relation to the positions of mathematical expressions. Different matrix calculations involve various mathematical operators, such as addition, multiplication, and determinants. The positions and functions of these symbols in mathematical expressions need to be accurately identified and positioned.

The extensibility of the code framework is crucial as technology continues to evolve and requirements change. During the system design phase, adopting a modular development approach ensures minimal dependencies between functional modules, making it easier to modify and upgrade these modules later on. This also facilitates rapid updates and real-time collaboration for front-end and back-end interfaces, as well as the expansion of functionalities.

## 1.3Work on this paper

To address the aforementioned issues, this project integrates artificial intelligence and visual animation technology to propose an innovative human-computer interaction method. The system, by incorporating YOLO-Mamba[2] image recognition technology and the Manim animation engine, can accurately identify handwritten matrices uploaded by students and present the entire computational process through dynamic animations. Unlike traditional mathematics teaching tools, this system not only focuses on displaying the results but also emphasizes the process, using clear visual derivations to help students understand the logic and principles behind each computational step.

Image Recognition: This project employs the YOLO-Mamba object detection algorithm, which, through an optimized training dataset, enhances the accuracy of recognizing handwritten matrices. The YOLO algorithm, by learning from multi-layer convolutional neural networks, can effectively identify each element in the matrix, overcoming issues such as variations in handwriting, noise interference, and image blurriness, ensuring high-precision recognition results.

The automated generation of animation processes allows the system to dynamically create corresponding computational process animations based on the matrix size, type, and required mathematical operations input by users, without human intervention. This mechanism is built on the modular sub-components and unified data interfaces of the Manim animation engine, enabling the system to achieve high flexibility and scalability. Whether it's matrix operations of different scales or various row-column determinants, addition, multiplication, and other operations, the system can automatically generate the corresponding animation effects, significantly enhancing user experience and interactivity.

Maintainability and Scalability To ensure the maintainability and scalability of the system, this project adopted a modular development approach, ensuring the independence and extensibility of each functional module. Through unified data interfaces and templated structures, the system can easily expand new mathematical operation functions and quickly update and maintain them in later stages. In particular, the design of the system ensures that new features can be rapidly deployed and supports multi-tasking and multi-scenario applications, enabling the system to meet evolving educational needs. All work in this document has been uploaded to https://github.com/rainbowyuyu/animate_cal, and all tasks can be reproduced according to the README instructions, with requests available for expansion and maintenance. The visualization platform address is https://animatecal-aesrxwe852bslylhgvrfxx.streamlit.app/, and the front end is updated in real-time through accessing the GitHub stream.

# 2 related work

## 2.1 hand-written numeral recognition

Handwritten digit recognition technology has made significant progress, especially in the development of deep learning, where methods based on Convolutional Neural Networks (CNNs) [3] have gradually become mainstream. CNNs can efficiently extract features from images through multiple layers of convolution and pooling operations, which is particularly advantageous when dealing with diverse characters and complex backgrounds, compared to traditional Optical Character Recognition (OCR) [4] methods such as edge detection, color analysis, and template matching. Although they have achieved good results on standard datasets, they often require longer training times and computational resources. Therefore, adopting more efficient CNN architectures, especially for handling noisy and complex background handwritten digits, remains a direction worth exploring.

## 2.2Corner detection algorithm

Corner detection is a key task in computer vision, widely applied in image matching, motion tracking, and other fields. The Harris corner detection algorithm [5] evaluates the gray level changes of local windows through a self-correlation matrix to determine whether corners exist in the image. The Shi-Tomasi algorithm [6], an improvement over the Harris algorithm, uses local brightness gradient information and structural matrix eigenvalues for corner detection, which better resists noise and enhances robustness. Despite its stronger adaptability, it has higher computational complexity and requires large-scale labeled data for training.

## 2.3YOLO neural network

YOLO algorithm is a real-time object detection method based on Convolutional Neural Networks (CNN). The core innovation of YOLO lies in transforming the object detection task into a regression problem, directly outputting the class of the object and its bounding box coordinates through a single forward pass of the entire image. Unlike traditional object detection methods, YOLO can effectively capture small objects in images with low spatial resolution and reduce the likelihood of background misidentification. However, for tasks like handwritten equations that require capturing details and contextual information from the image, as well as adding attention recognition, improvements are still needed.

## 2.4Mamba backbone network

Mamba backbone network is a convolutional neural network architecture proposed in recent years for visual tasks, aiming to improve the performance of traditional convolutional networks in complex visual tasks. Compared with other backbone networks, Mamba network introduces dynamic convolution and adaptive feature fusion mechanisms, enabling it to more effectively capture image details and contextual information.

# 3 Data collection and annotation

## 3.1The raw data set

The original data set "Handwritten math symbols dataset" has 82 classes, including their own operation symbols and numbers and letters. Each class contains about 10,000 black and white images in jpg format of 45x45. The url of the data set is https://www.kaggle.com/datasets/xainano/handwrittenmathsymbols/data.

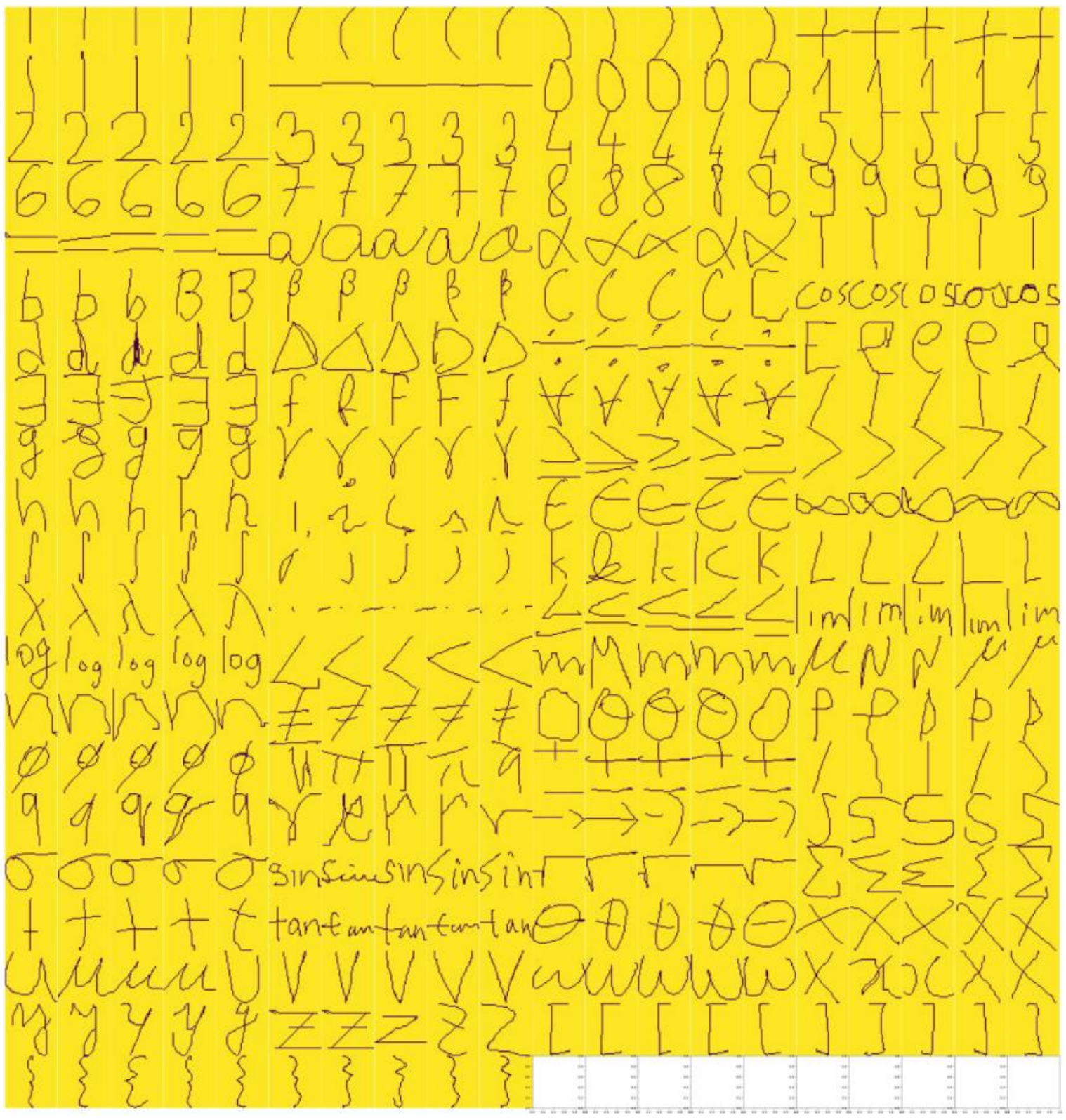

Figure 3.1 Sample of the original data set

## 3.2Data set generation

The data set generation adopts the operations of adding alpha channel, image transverse and longitudinal transformation, adding noise and so on in digital image processing. Multiple versions of iterations and attempts are carried out, and finally the automatic data set generation and automatic annotation functions are realized.

The version iteration of the data set generator is also an important basis for the model version iteration. The basis for the version performance iteration when the model is introduced later is the version of the data set.

The automatic generation and annotation of data sets solve the huge workload of manual writing data sets and labelimg annotation.

Table 3.1 Summary of data set versions

| edition | function |
|---|---|
| v0 | Randomly arrange the numbers |
| v1 | Add an alpha channel |
| v2 | Generate matrix elements |
| v3 | generated matrix |
| v4 | Generate a stable handwritten matrix |

（1）v0

Read all files and build a list, randomly select an element and place it in a random position of random size. When generating annotations, change the width and height of the annotations with random values.

Problems caused by v0: The upper layer elements cover the lower layer, and the image in the label is not complete, resulting in very serious overfitting.

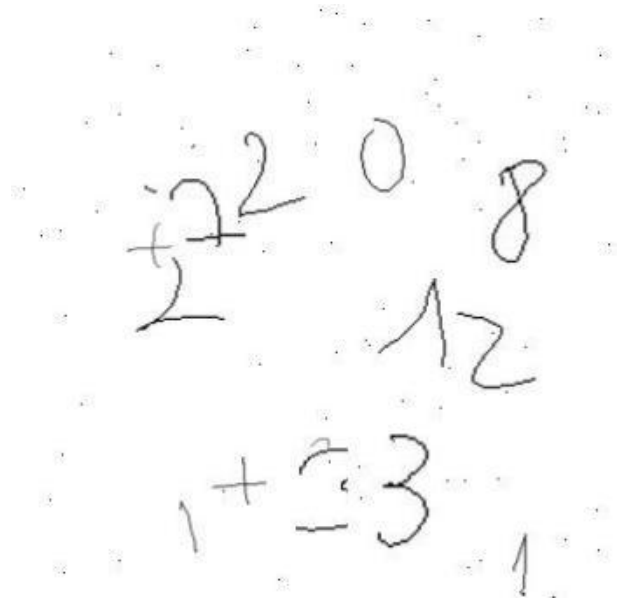

Figure 3.2 Sample images generated by v0

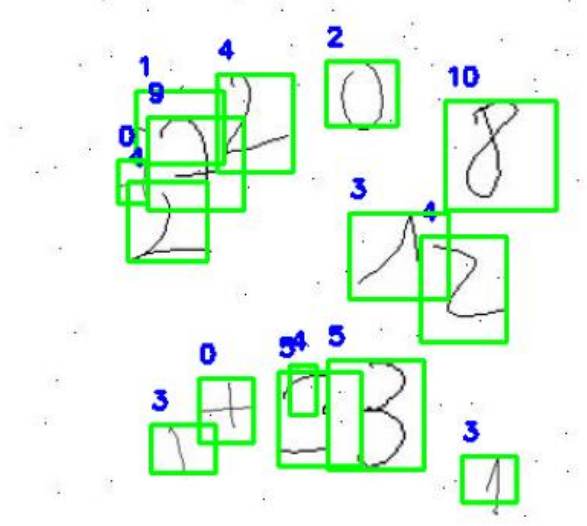

Figure 3.3 Sample annotated by v0

Note: The labels in the image are the categories of the image, and their classes mapping relationship is: [{0: +}, {1: -}, {2: 0}, {3: 1}, {4: 2}, {5: 3}, {6: 4}, {7: 5}, {8: 6}, {9: 7}, {10:8}, {11:9}, {12: =}, {13: [}, {14:]}, {15: times}], all labels in all figures in section 3.1.2 are presented according to this mapping relationship.

（2）v1

The following adjustments have been made to address the problems encountered in v0:

-The number of elements is reduced, and the alpha channel is added to the original data set to become a transparent grayscale image of png. In this way, there will be no white background of the upper image covering the lower image.

-The generated image is no longer a fixed square image, but the total size of the image is changed appropriately at random to reduce overfitting.

Problems caused by v1: Multiple elements overlap and cover each other. The elements in the two labels cannot be distinguished when they overlap, resulting in serious overfitting.

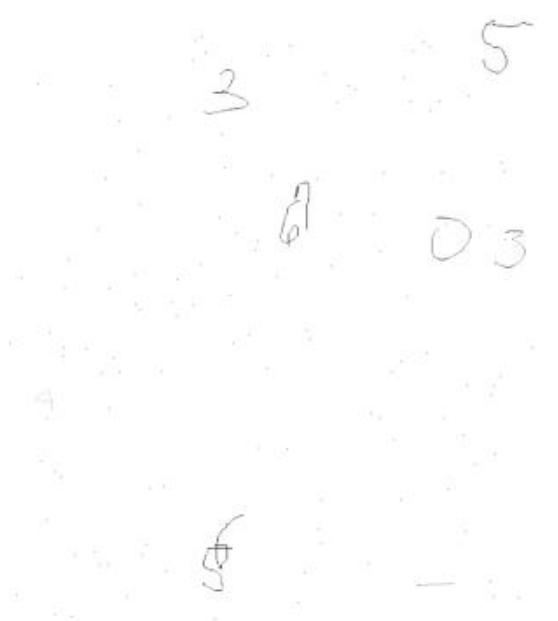

Figure 3.4 Sample images generated by v1

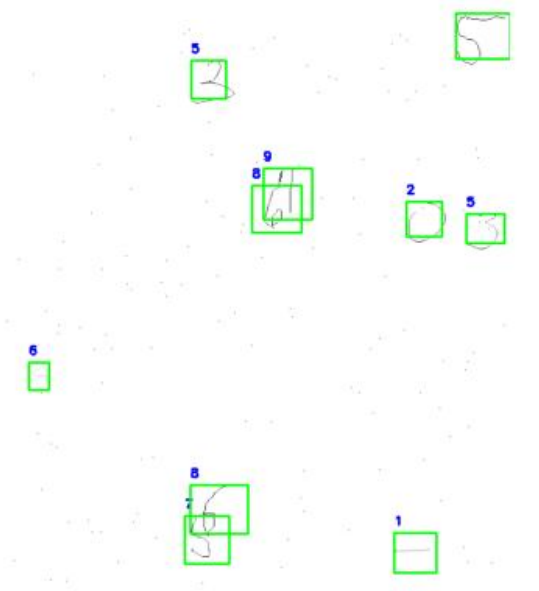

Figure 3.5 Sample annotated by v1

（3）v2

The following adjustments have been made to address the problems encountered in v1:

-Change the logic of random point generation. Instead of directly adding images from the list and placing them randomly, n points are generated randomly first to ensure the interval between these n points, and then these n points are positioned as the upper left point of the image to avoid repeated images.

-Rewrote the alpha channel generation method to reduce the threshold for removing white backgrounds and prevent elements from having too few features.

V2 has been able to generate basic multi-element data sets stably, and continue to generate matrices and elements in the matrices, so that features are more diversified and generalized. This is achieved by adaptive splicing after generating matrix elements.

Generation of matrix elements: According to the actual situation, 1-3 digits are generated. When the number of bits is>2, the first digit has a probability of being negative, and the size and interval of appropriate random numbers are arranged.

Generation matrix: An algorithm for adjusting the size of images according to the number of adaptive elements and an algorithm for placing matrix elements according to the number of adaptive rows and columns are written.

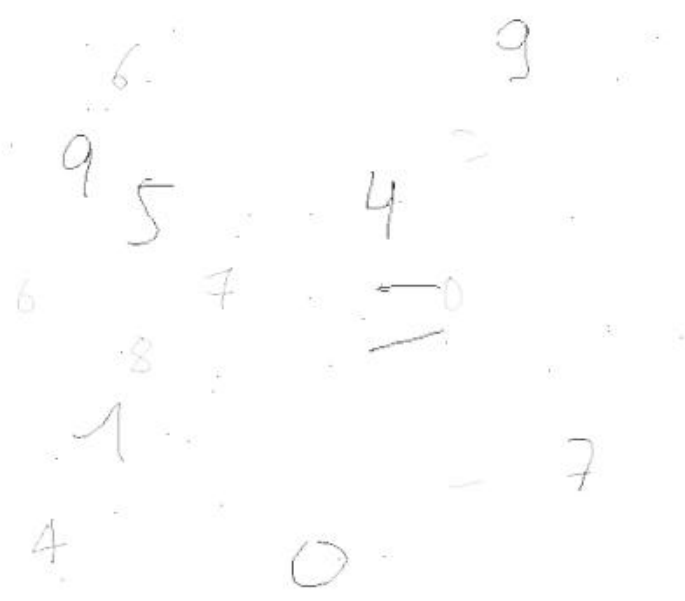

Figure 3.6 Example of alpha pattern generated by v2

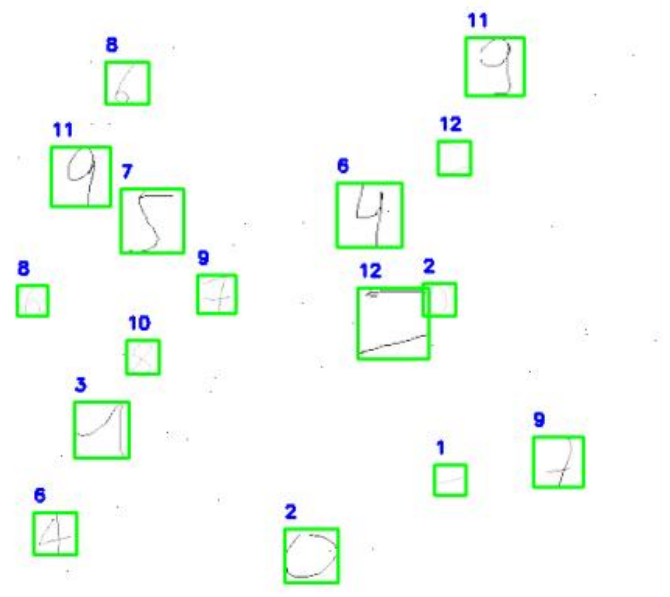

Figure 3.7 Sample alpha annotation generated by v2

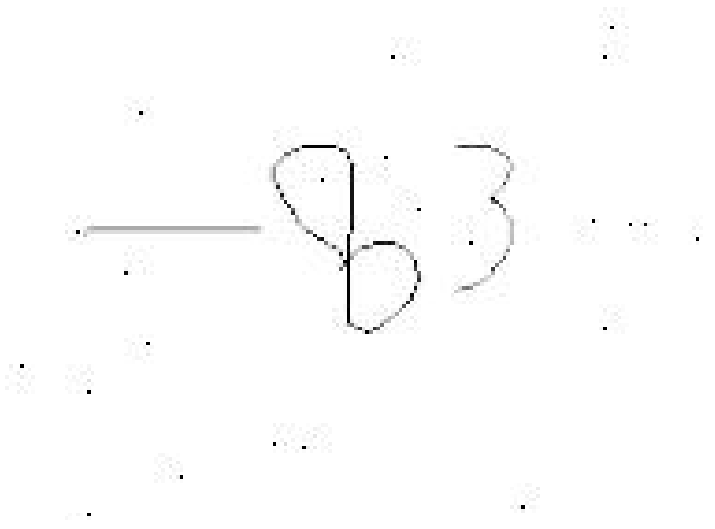

Figure 3.8 Matrix element pattern generated by v2

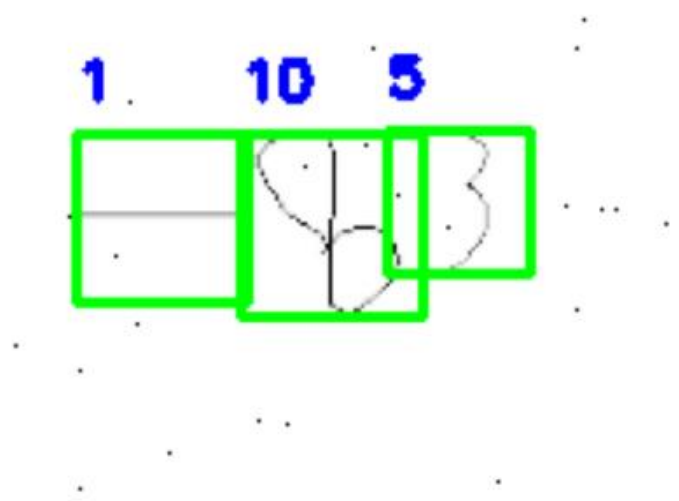

Figure 3.9 Sample matrix element labeling generated by v2

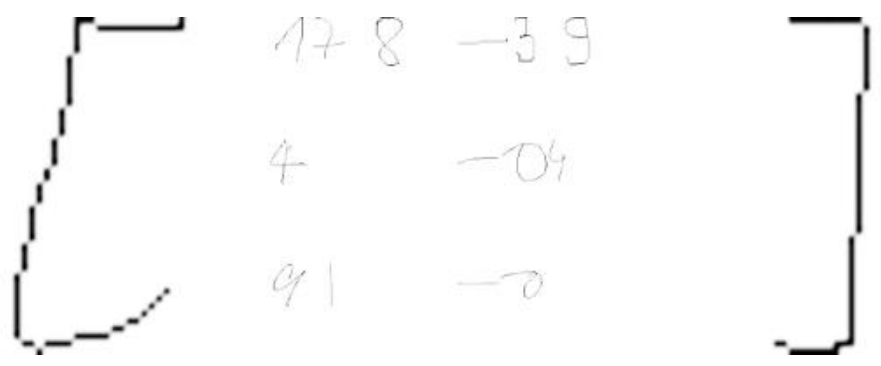

Figure 3.10 Matrix pattern generated by v2

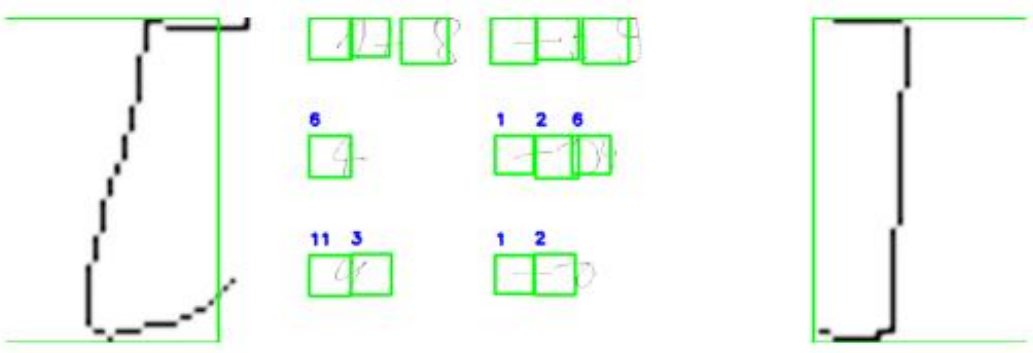

Figure 3.11 Sample matrix annotations generated by v2

（4）v3

The following adjustments have been made to address the problems encountered in version v2:

-Fix the misalignment of labels in version v2.

-Adjust the adaptive vertical stretching relationship of the left and right brackets of the matrix, and randomly move up, down, left and right to mimic the matrix written by hand.

-An adaptive centering algorithm for elements in the matrix is added to prepare for subsequent clustering and segmentation, which more imitates the characteristics of handwritten matrices.

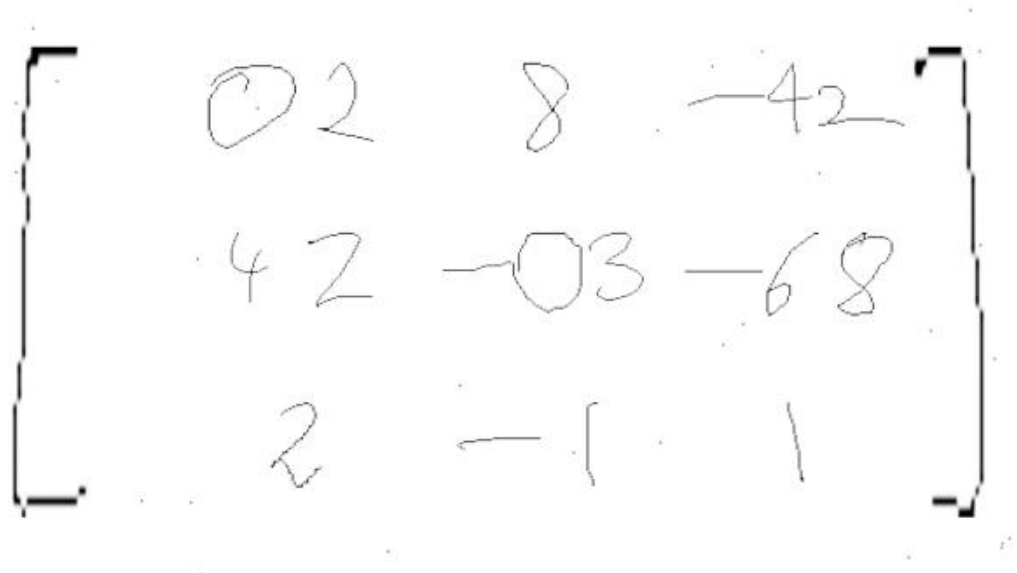

Figure 3.12 Matrix pattern generated by v3

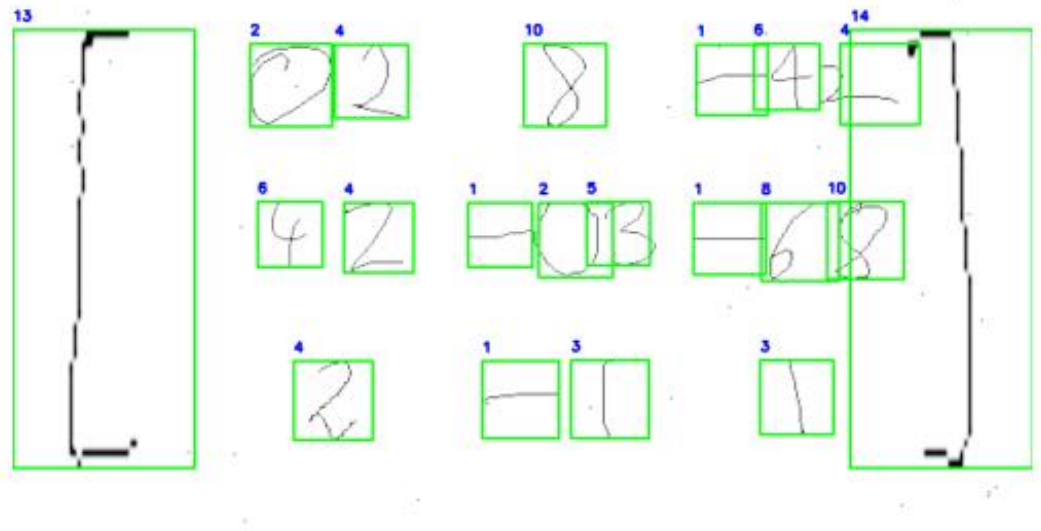

Figure 3.13 Matrix annotation sample generated by v3

（5）v4

The following adjustments have been made to address the problems encountered in v3:

-Adjust the thickness of elements in the matrix, adjust the generation logic of alpha elements, and add the alpha channel when adding elements to the matrix.

-Adjust the noise generation logic. The noise is now gradually decreasing from the center to the surrounding alpha value, mimicking the actual image preprocessing noise.

-The matrix adaptive position centering, row and column positions, spacing and other features have been optimized. Now the matrix is very close to the handwritten matrix.

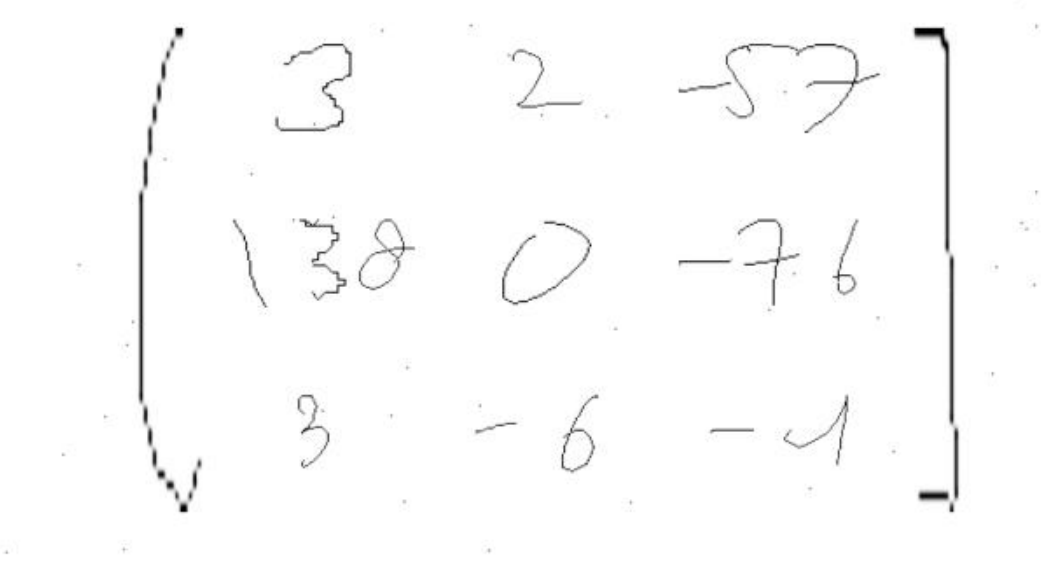

Figure 3.14 Matrix pattern generated by v4

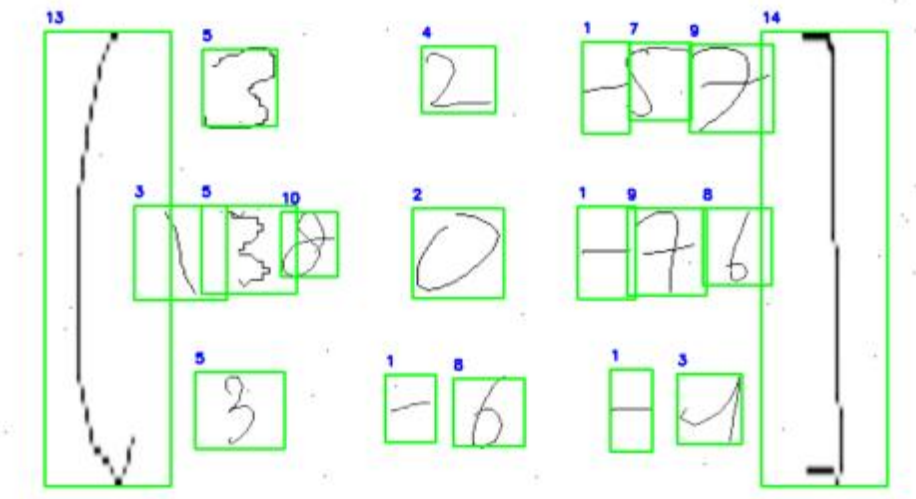

Figure 3.15 Example of matrix annotation generated by v4

Through the optimization and adjustment of the matrix generation algorithm by multiple iterations, the final stable version of the matrix automatic generation and annotation data set was obtained, and the model training was entered, with a total of 300,000 images. The ratio of the training set to the test set was 9:1, and the sample of the data set is shown in Figure 3.16.

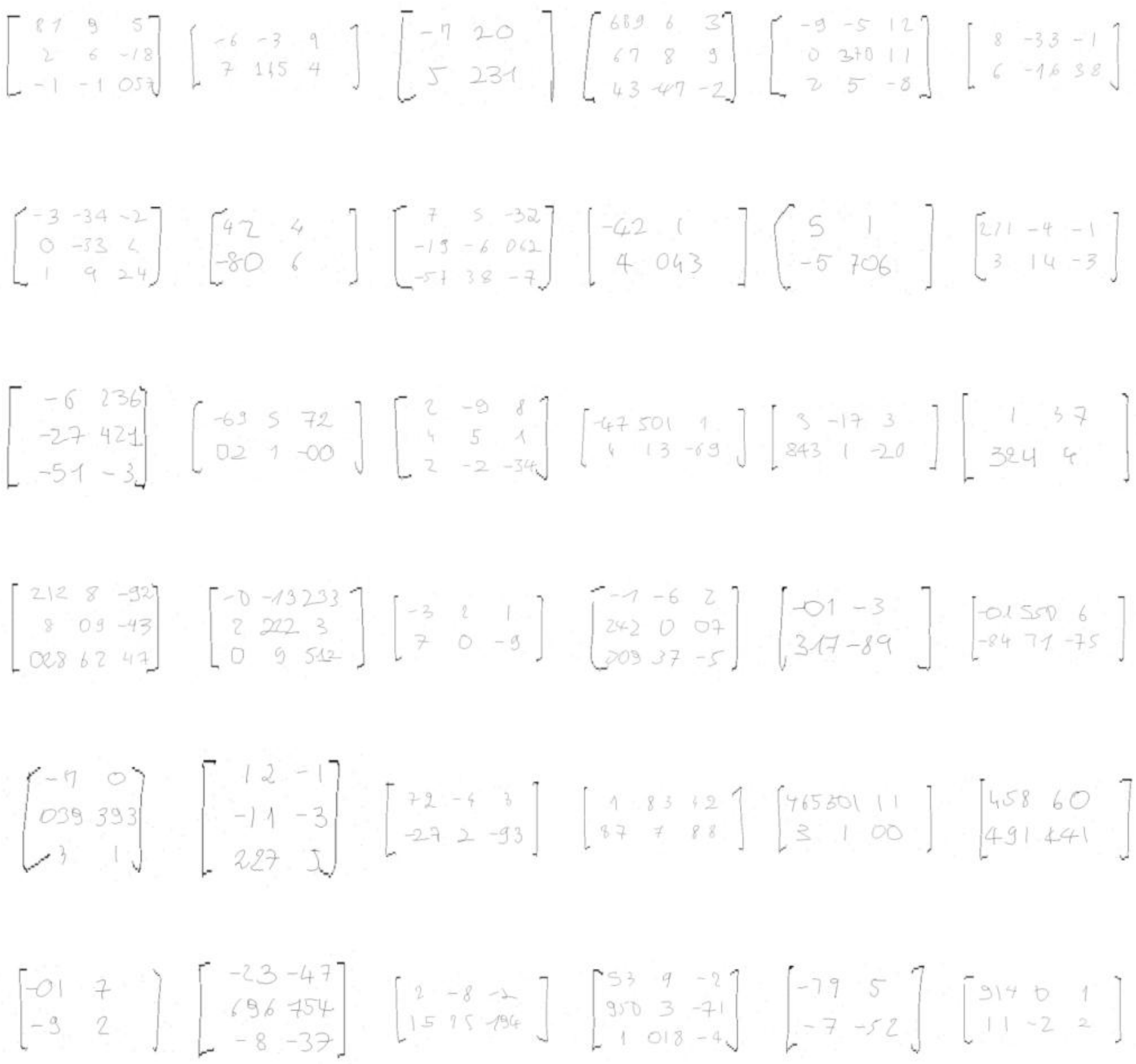

Figure 3.16 Sample graph of generated data set

The algorithm flowchart of the final stable version is shown in Figure 3.17, and the specific code is shown in Appendix I Algorithm 1.

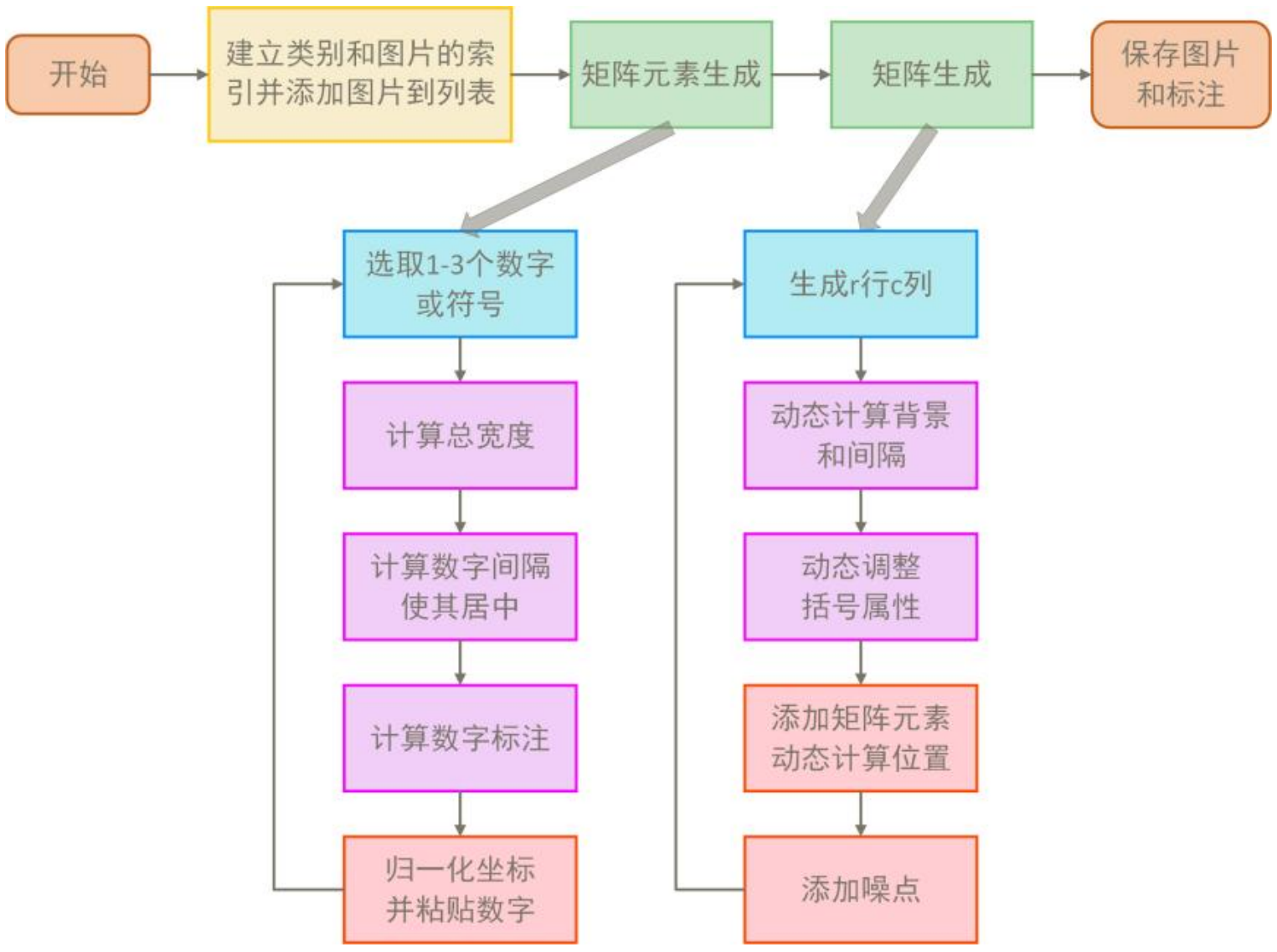

Figure 3.17 Flowchart of the data generation algorithm

# 4 Key methods

## 4.1Mathematical equation detection

### 4.1.1 Mathematical equation model training

In terms of network architecture and model training, this project selected the lightweight YOLOv11n as the basic object detection framework. It performed dual optimization at both structural and mechanism levels based on the spatial distribution characteristics of handwritten equation images. Due to the integration of YOLO, Mamba, and CoordAttention modules, I refer to the network as YMCANet in the following text. The network architecture is divided into three major modules: Backbone, Neck, and Head.

Layer Backbone builds upon the existing C2F structure by incorporating a more expressive C3K2 multi-scale convolution module, significantly enhancing the ability to capture complex edges and stroke features. To further boost network performance, a Mamba backbone network is introduced, improving the overall expressive power and computational efficiency of the model. Additionally, VSS, VCM, and XSS modules are added to the network, effectively enhancing the fusion capability of multi-scale information and the accuracy of feature extraction.

In terms of attention mechanism, the original PSA module is optimized and replaced with CoordAttention coordinate attention mechanism[7], so that the network can not only pay attention to content information when recognizing the matrix, but also grasp the spatial position of characters in the image more accurately, thus improving the recognition accuracy and robustness.

（1）VSS Visual State Space Module

The VSS module effectively models visual elements in images by introducing a visual state space. In traditional visual recognition tasks, information in images is typically static. However, in VSS, each visual element is treated as a dynamic state that can change during image processing. This module captures dynamic information at different levels of visual data, such as changes in objects, background variations, and the relationships between visual elements.

（2）VCM convolution feature fusion module

The VCM module is primarily used to integrate feature information from different convolutional layers, enhancing the network's ability to understand complex image features. It achieves this by weighting and fusing the output features from multiple convolutional layers, ensuring that the network can effectively combine information at different levels, thus increasing the diversity and depth of feature representation. VCM enables the network to better handle image information with varying scales and resolutions, thereby improving the accuracy of feature extraction, especially when dealing with complex scenes or images.

（3）The XSS extension receptive field module

The XSS module, by expanding the receptive field of convolutional kernels, can capture a broader range of contextual information. This module leverages multi-layer convolution operations to effectively extend the network's perception of distant pixels in images, making the model more accurate when processing features over large areas. The XSS module has strong capabilities in capturing long-range dependencies and complex backgrounds in images, enhancing the network's performance in multi-scale feature fusion, especially in scenarios with high image resolution or complex structures.

（4）CoordAttention Coordinate attention mechanism

CoordAttention is an attention mechanism based on spatial coordinates. Unlike traditional channel attention mechanisms, it not only focuses on content information in the image but also introduces coordinate information to accurately grasp the spatial position of characters within the image. By weighting the coordinates, CoordAttention can make the network more focused on key areas, enhancing its ability to understand spatial structure and layout. When recognizing matrices, CoordAttention can accurately locate the specific positions and shapes of characters, thereby improving recognition accuracy and robustness, especially in complex backgrounds or when characters are distorted.

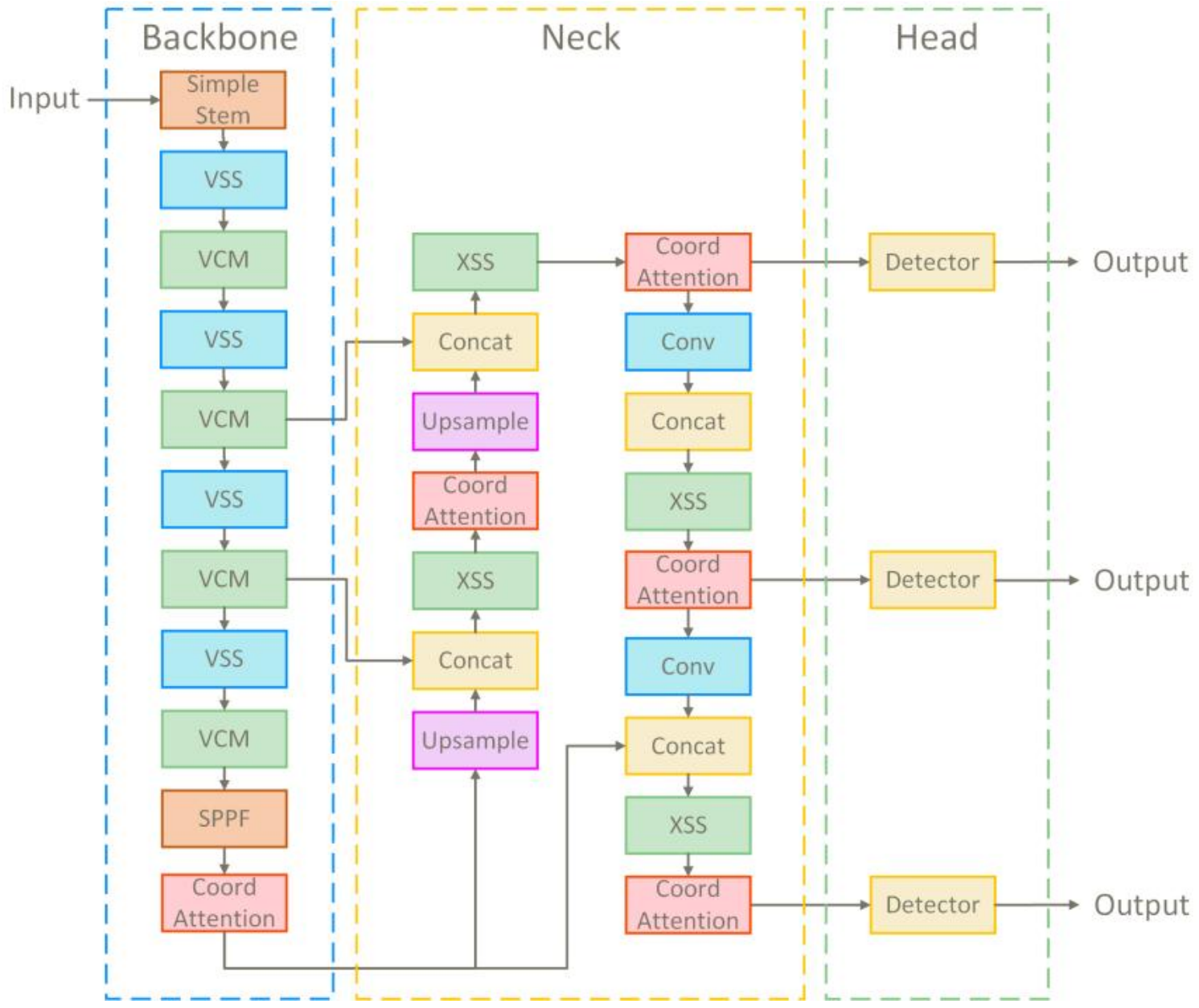

Figure 4.1 YMCANet Network model diagram

CoordAttention The coordinate attention mechanism can capture the spatial relationship of a specific location and make use of it in the attention calculation. The algorithm flow chart is shown in Figure 3.6.

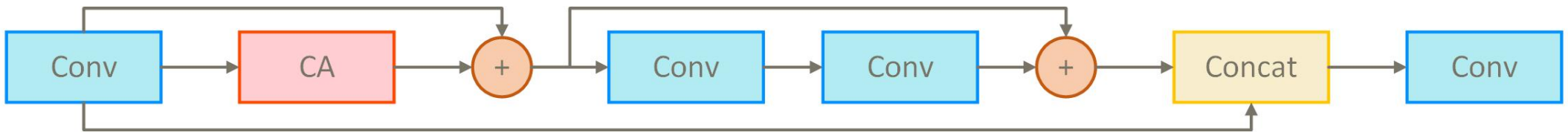

Figure 4.2 CoordAttention Module algorithm flow chart

Algorithm calculation formula:

$$\text{output} = \text{Conv1} \times 1(\text{concat}(a\,, + \text{Cooratt}(b) + \text{FFN}(b))) \quad (1)$$

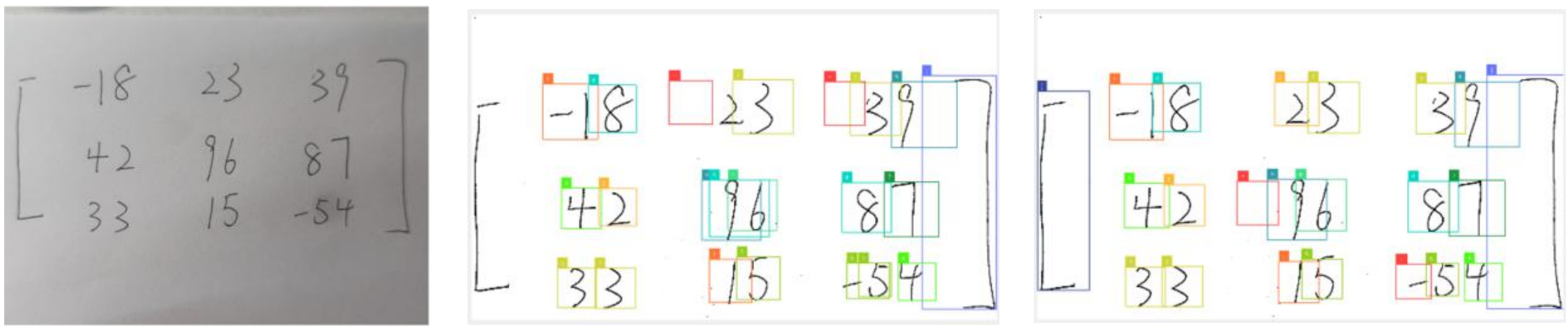

artwork master PSA module CA module

Figure 4.3 Comparison of attention module effects

The training phase utilized a self-developed handwriting matrix dataset (comprising 300,000 images with various arrangements and interference forms). Within 10 rounds of training, the model achieved an accuracy above maP500.95, demonstrating excellent convergence speed and generalization ability. Additionally, the training process involved pre-training transfer between multiple generations of models, combined with fine-tuning and hierarchical visualization, effectively enhancing the model's recognition stability under complex writing styles, providing a solid foundation for subsequent step-by-step visual computation.

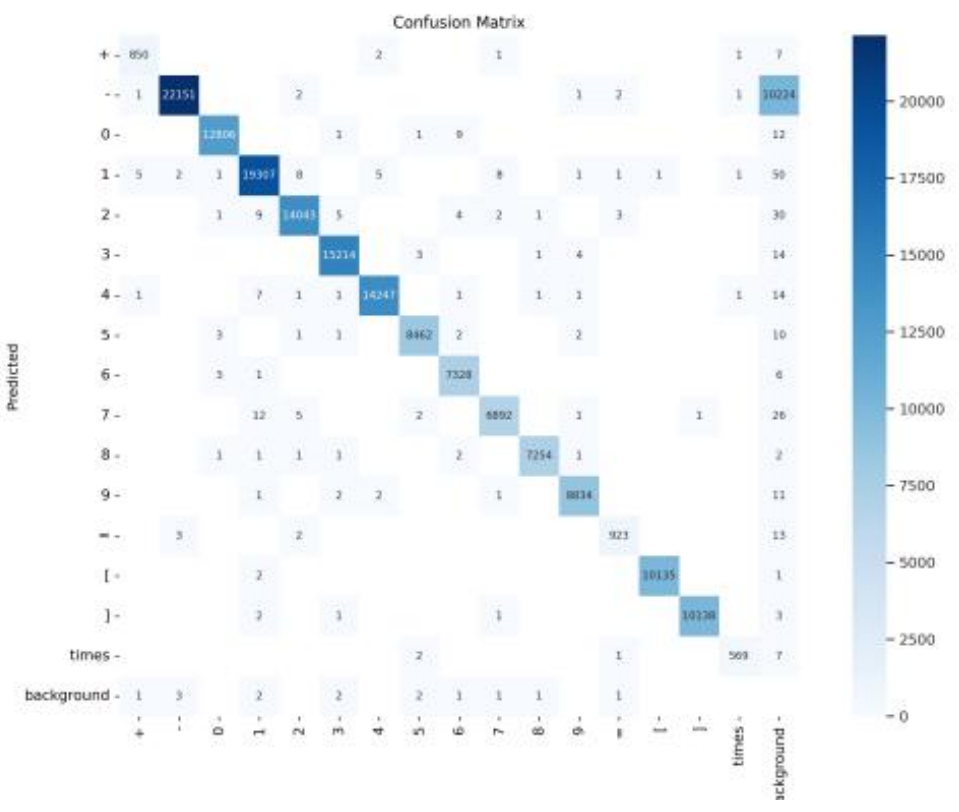

Figure 4.4 Matrix pattern generated by v4

F

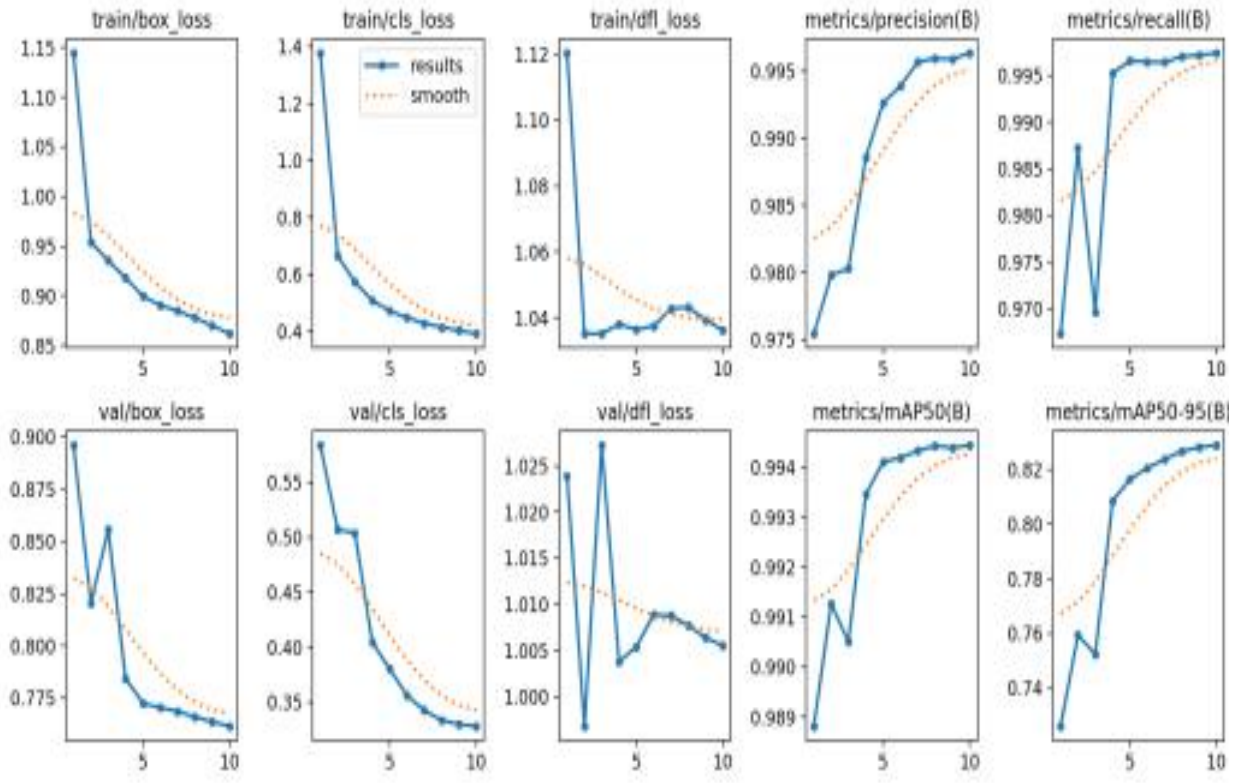

igure 4.5 Examples of matrix annotations generated by v4

## 4.1.2 Formula image preprocessing

Since the generated data set is binary, it is necessary to carry out image preprocessing algorithm and appropriate scaling preprocessing for the input captured image. The specific algorithm flow chart is shown in Figure 3.10.

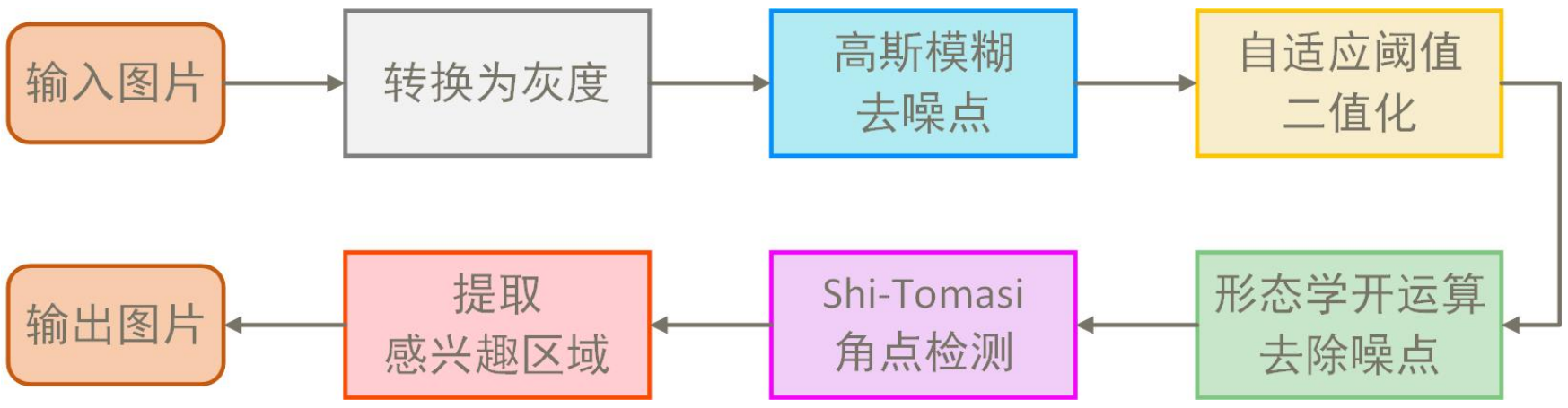

Figure 4.6 Flow chart of image preprocessing algorithm

(1) Convert to grayscale image and Gaussian blur denoising:

Application cv2.GaussianBlur smooths the image. By using a 5x5 Gaussian kernel through convolution operations, the image is blurred, with the standard deviation automatically calculated based on the kernel size. This process can smooth the image, remove fine noise, making subsequent thresholding and morphological processing more accurate and reliable. The specific formula for the operation is as follows:

$$I_{\text{blurred}}(x,y)=\sum_{i=-k}^{k}\sum_{j=-k}^{k}I(x+i,y+j)K(i,j) \tag{2}$$

(2) Adaptive threshold processing:

Binary image processing is used to retain only black and white pixels, making it easier to distinguish between foreground and background. The cv2.adaptiveThreshold method is employed, automatically adjusting the threshold based on the brightness of the surrounding area of each pixel, effectively addressing uneven lighting conditions. The cv2.ADAPTIVE_THRESH_GAUSSIAN_C method is utilized, where the weighted average of pixels within a neighborhood serves as the local threshold, thus adapting to images under various lighting conditions. The specific formula for adaptive threshold calculation is as follows:

$$T(x,y)=\frac{1}{N}\sum_{i=-\left\lfloor\frac{N}{2}\right\rfloor}^{\left\lfloor\frac{N}{2}\right\rfloor}\sum_{j=-\left\lfloor\frac{N}{2}\right\rfloor}^{\left\lfloor\frac{N}{2}\right\rfloor}I(x+i,y+j) \tag{3}$$

(3) Morphological open operation to remove noise:

Clean up small noise points in the image and enhance the connectivity of the target area. An opening operation was used, first erosion followed by dilation, to remove small white noise points. The cv2.morphologyEx function with a cv2.MORPH_RECT rectangular structuring element was applied to the image to improve the effectiveness of subsequent corner detection. The specific formula for the operation is as follows.

Corrosion operation:

$$E(x,y)=\min_{(i,j)\in S}I\,(x+i,y+j) \tag{4}$$

expansive working:

$$D(x,y)=\max_{(i,j)\in S}E\,(x+i,y+j) \tag{5}$$

(4) Shi-Tomasi corner detection:

The Shi-Tomasi corner detection algorithm cv2.goodFeaturesToTrack was used to detect areas of intense change in images, improving accuracy compared to commonly used Harris and Fast clustering methods, and adding the selection of optimal corners. The maximum number of corners extracted is limited to 100, with higher quality corners being selected. Corner detection provides positional information for target areas, aiding subsequent steps in more accurately extracting regions of interest. The formula for the gradient covariance matrix of each pixel's neighborhood is as follows:

$$M = \begin{bmatrix} \sum I_x^2 & \sum I_x I_y \\ \sum I_x I_y & \sum I_y^2 \end{bmatrix} \tag{6}$$

(5) Boundary range calculation and interest area extraction and transformation:

The minimum and maximum boundaries of the region of interest are calculated from the extracted corner coordinates, and the edge offset is further ensured that these boundaries do not exceed the actual range of the image. The extracted slices are returned as a color image.

The final results of preprocessing are shown in Figure 4.7.

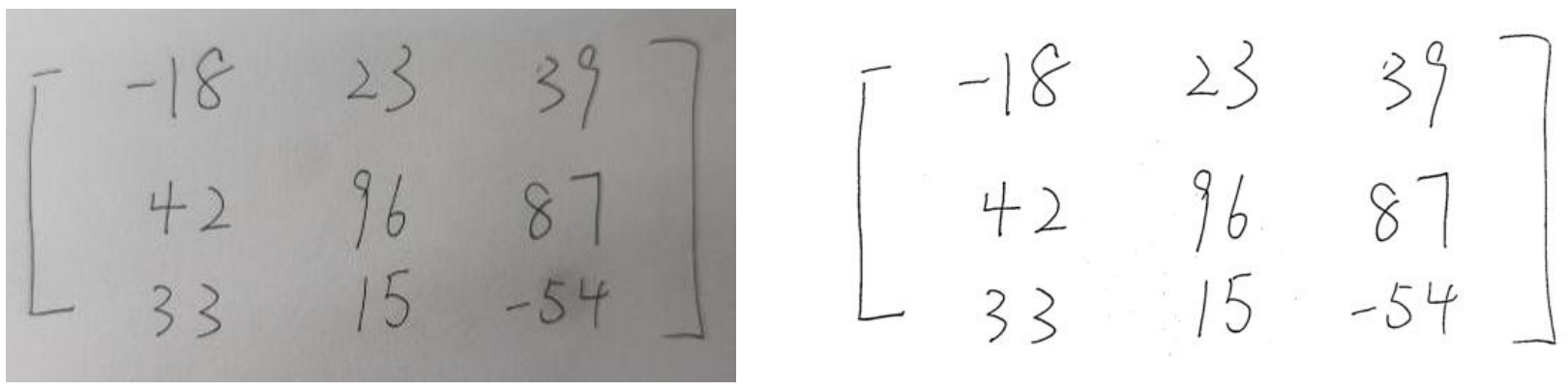

artwork master Preprocessing results

Figure 4.7 Preprocessing results

## 4.1.3 Formula content detection

In the content detection module of the algorithm, the system calls the YOLOv11 model provided by ultralytics and uses the supervision library to uniformly monitor and process the detection results. It mainly returns the center coordinates, width and height, as well as category annotations for each recognition box. To improve recognition accuracy and avoid duplicate recognition issues, the system introduces a deduplication mask mechanism based on IoU (Intersection over Union). By calculating the overlap ratio between two detection boxes, it filters out redundant detection results with low confidence, effectively reducing the risk of overfitting. Additionally, the system generates masks for each element, providing spatial foundations for subsequent DBSCAN clustering segmentation and matrix structure reconstruction. This module not only ensures precise localization of handwritten characters but also lays a solid data foundation for the structural recognition and visual computation of the entire matrix.

The system also adds a return value to return all the annotated masks, which is ready for subsequent clustering segmentation. The processing results are shown in Figure 4.8.

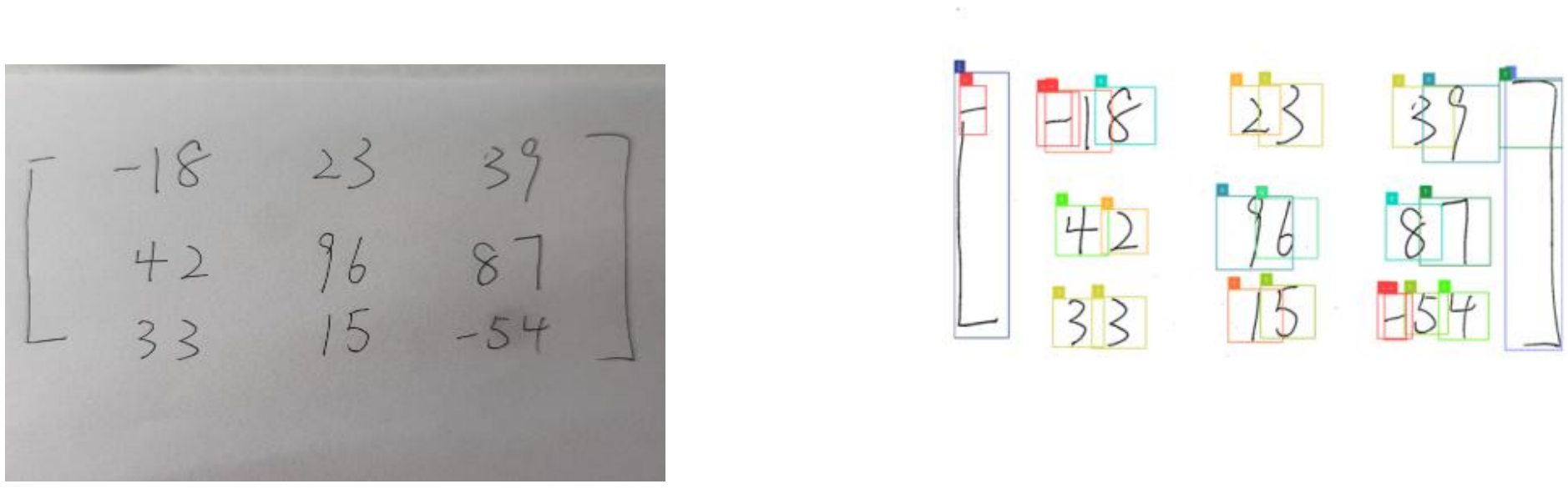

artwork master　　Preprocessing results

Figure 4.8 Yolo detection results

### 4.1.4 Mathematical formula number extraction

After receiving the return value of 4.1.3 detection, cluster and segment the matrix to determine its row and column numbers. Then draw the segmentation lines to prepare for determining which area each individual number symbol belongs to in the future. Finally, return a list of horizontal and vertical segmentation lines. The algorithm flowchart is shown in Figure 4.9.

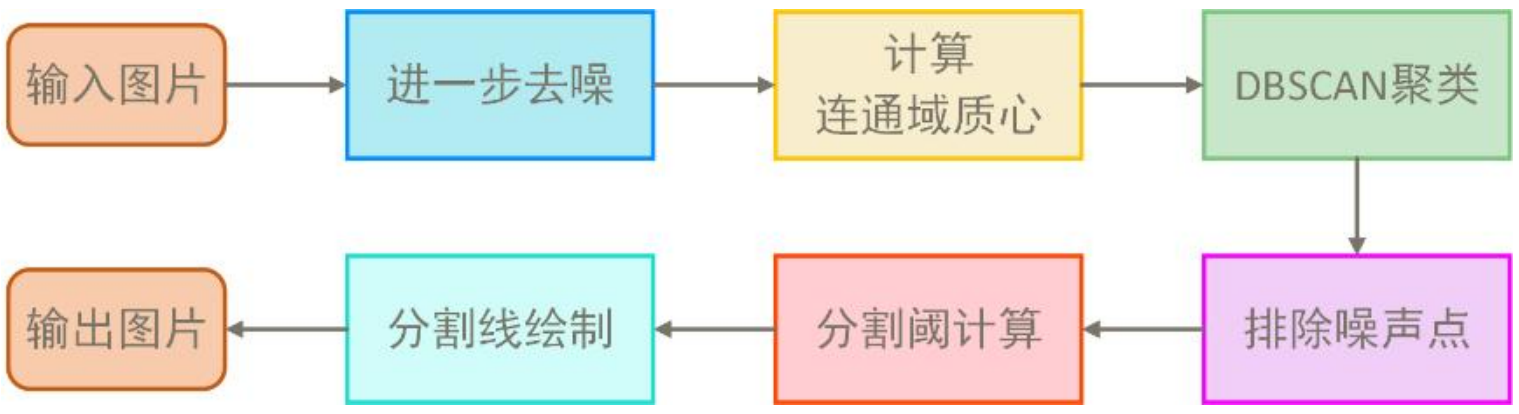

Figure 4.9 Flow chart of clustering segmentation algorithm

Using cv2.findContours to find connected regions in images, the centroid of these regions is calculated through cv2.moments image moments. Given a point p and radius $\epsilon$, if there are enough neighboring points within this radius, these points are considered to belong to the $(cX, cY)$ same cluster. Based on the number of centroids, the initial calculation of the row and column numbers of the region grid is performed. For each centroid, its associated rectangular area is calculated. Multiple segmentation lines are drawn using the intersection points of clustered regions, and then the segmentation lines are filtered using deduplication operations.

The line validation checks whether there is a centroid on both sides of a certain line. If not, this line is excluded to ensure that each line indeed serves as a row or column division in the matrix. Ultimately, two lists are obtained: horizontal and vertical lines, which are then returned to the number extraction module. The result is shown in Figure 4.10.

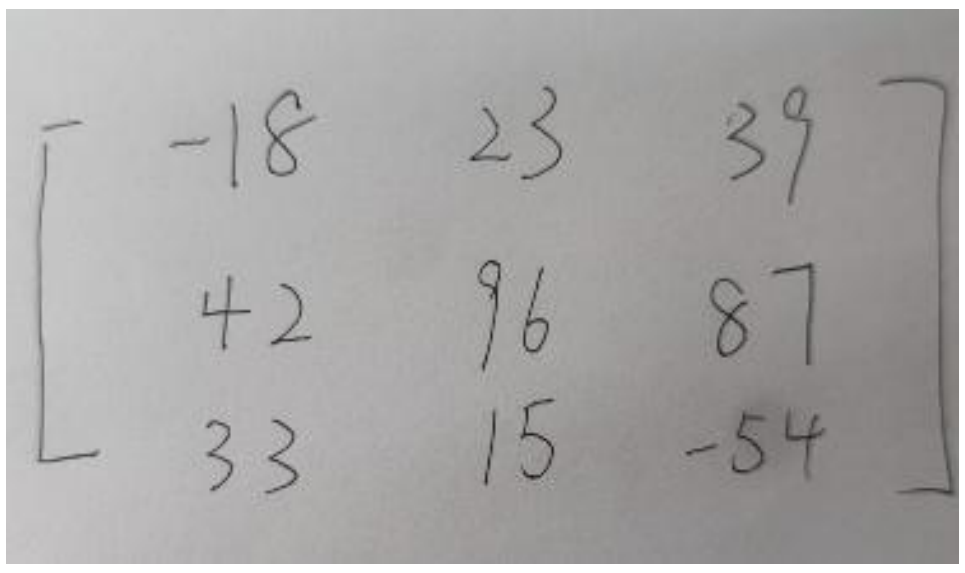

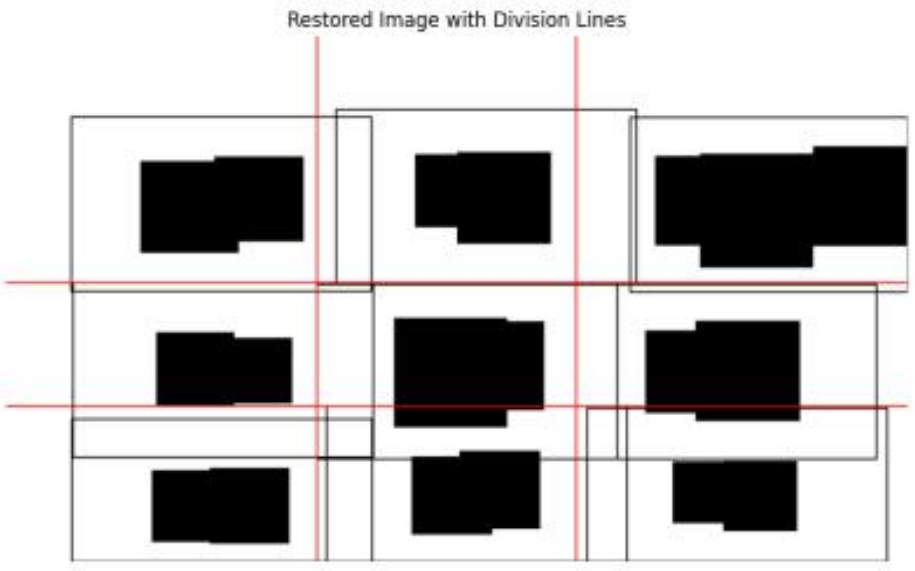

artwork master Preprocessing results

Figure 4.10 Results of clustering segmentation algorithm

First, through normalization operations, we determine the exact relative positions of each element. To ensure that the original positional relationships remain unchanged when placing corresponding elements in each row and column, we sort the x_center values of each element and add them to a list. Traverse the entire list, and based on the relative positional relationships between elements x_center and y_center with respect to the partition lines returned by the clustering segmentation algorithm in Figure 4.10, place the elements in their corresponding positions. After concatenating the strings, perform type conversion to return an array format of numpy for easier Latex presentation later. The result is shown in 4.11.

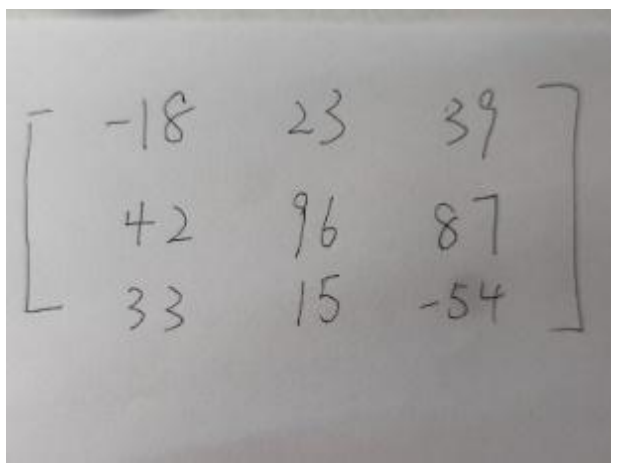

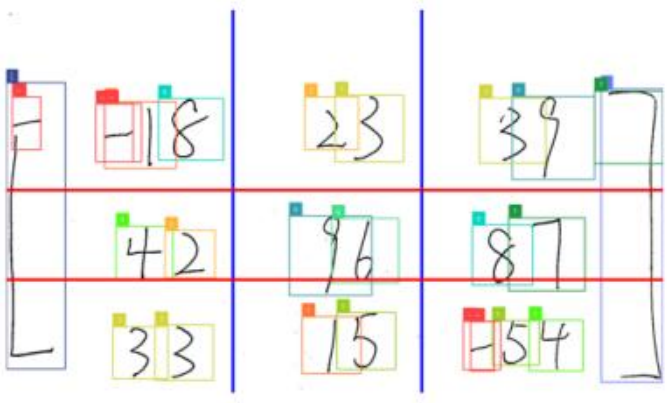

$$\begin{bmatrix} -18 & 23 & 39 \\ 42 & 96 & 87 \\ 33 & 15 & -54 \end{bmatrix}$$

artwork master Split line results Latex matrix display

Figure 4.11 Digital extraction results

## 4.2Step-by-step solution animation module

In terms of animation module construction, the system carries out underlying graphics rendering based on the mathematical visualization engine Manim of Python, and completes frame-by-frame video output combined with ffmpeg, realizing dynamic visualization demonstration of matrix operation process.

By inheriting the basic graphic objects and text classes from Manim, a custom class SquTex was constructed. This class encapsulates the combination of squares and text for each matrix element, enabling absolute position control, format style adjustment, and animation presentation of individual elements. Subsequently, based on SquTex, a MatrixCal class was designed to achieve logical expression and layout control of two-dimensional matrix objects, supporting high-precision positioning and bracket labeling for each element. At the computational level, a MatrixDet class was developed for determinant calculations, supporting the extraction of multiplication paths from the main and secondary diagonals and calculating intermediate values. The encapsulated animation methods are called to display each group of multiplication and addition processes in a diagonal manner, highlighting key paths and the logic of result derivation. For matrix addition and multiplication, the MatrixMath class introduces a dual-input structure, using two original matrices as input objects for sequential operations. It achieves element-by-element alignment for addition and item-by-item expansion for multiplication through constructing progress vectors and calculating annotation texts. This enhances the transparency and interpretability of the calculation process. The entire animation system adopts a modular design, with clear inheritance and independent functional units, ensuring scalability and reusability. It is a core supporting component that professionally and dynamically presents abstract mathematical computation logic, as shown in Figure 4.12.

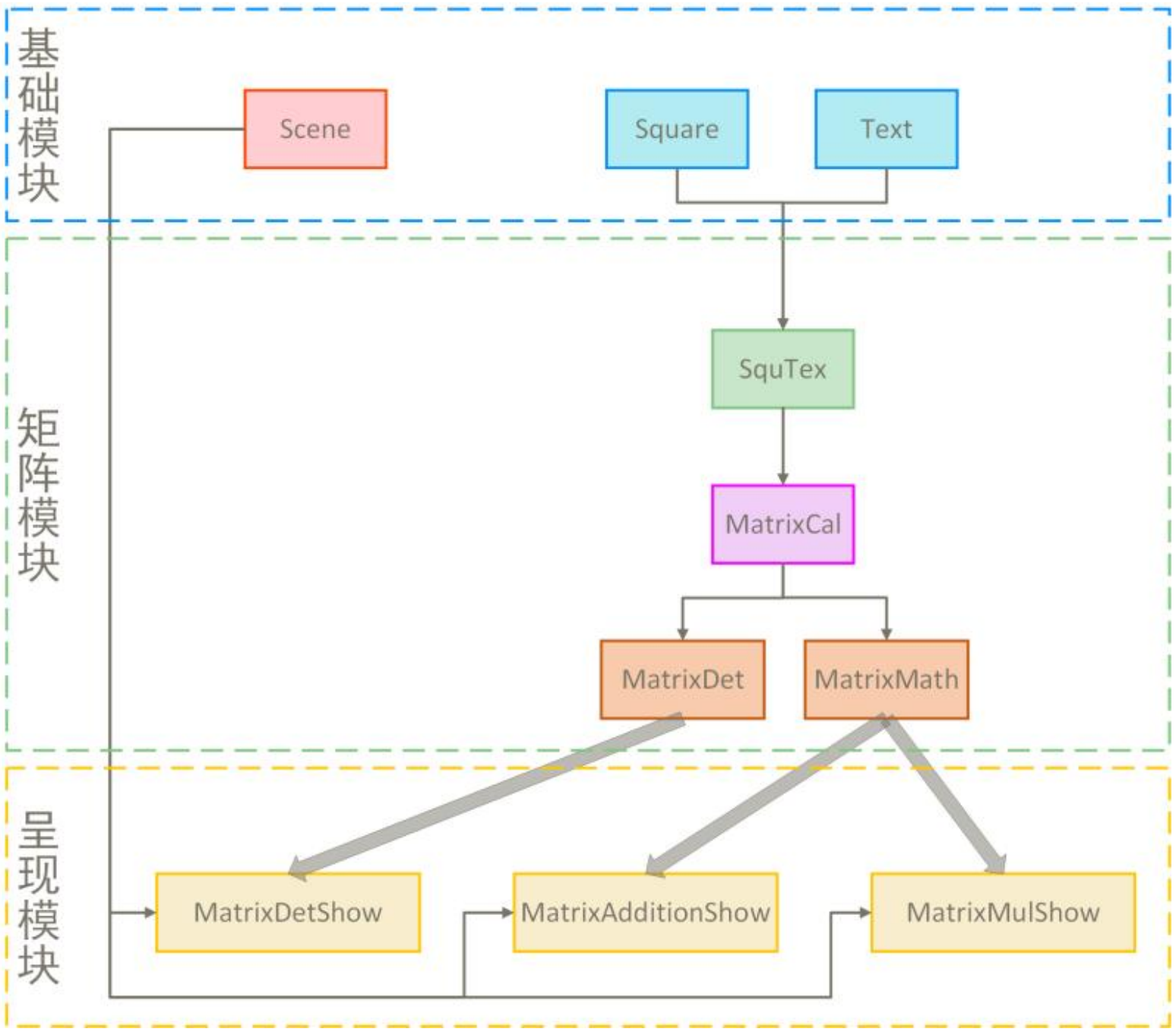

Figure 4.12 Animation module class object inheritance diagram

### 4.2.1 Step-by-step solution frame by frame rendering method

The Manim engine provides a complete vector animation scene management system by inheriting from Scene or ThreeDScene. This project uses the construct() function as the main entry point for animations, where graphic objects and animation control commands are gradually called to achieve dynamic computation demonstrations. All rendering results are generated at the frame level using vector graphics (SVG), and ultimately synthesized and encoded through ffmpeg for high-quality MP4 video output. This mechanism offers excellent resolution scalability and smooth animation, meeting the visualization requirements of highly complex mathematical expressions.

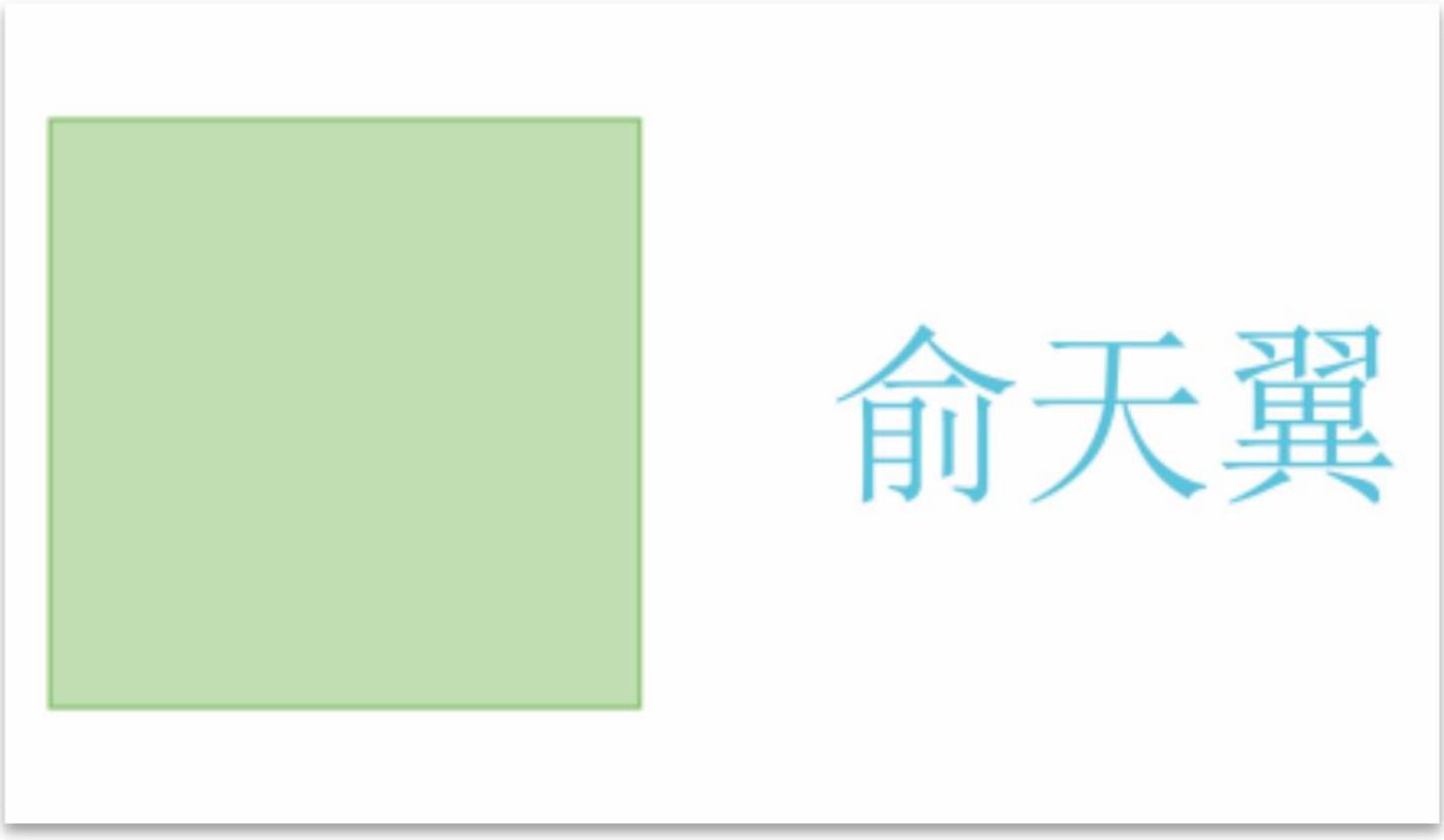

Figure 4.13 Example of a basic module

### 4.2.2 The control scheme for each animation element

By inheriting the and classes shown in the basic SquareTextmodule of Figure 2, define an element as a group of squares and text. When creating the structure, input a list or string as a variable parameter to create a list of square text combinations. The specific construction function is shown in Code 1.

Code 1 SquTex Constructor

```
Input  : tex: string or list, font: string, kwargs: additional arguments
         for Square
Output: Constructed object with arrangement of squares and text
self.tex ← tex;
Call parent constructor;
for i = 0 to len(tex) - 1 do
    Create a VGroup object;
    Add a Square object to VGroup with kwargs;
    Add a Text object to VGroup with the current element of tex and
     specified font;
    Add VGroup object to current object;
end
Arrange objects with zero spacing;
```

The SquTex is a fundamental graphic composite class in the animation system. By inheriting from the Manim's VGroup object, it models each mathematical element as a "square border + text" composite structure, featuring high-precision position management and style control capabilities. During the construction phase, it can accept a list of strings and automatically generate a set of orderly arranged graphic units. This class integrates methods such as position sorting, symbol addition, style modification, and unit animation grouping, supporting the positioning, highlighting, bolding, and style switching of any element in the matrix. Through encapsulation of animation interfaces like animate_one_by_one() and add_bracket(), it achieves frame-by-frame control and diverse visual effects during the matrix presentation process. The specific effects are shown in Figure 4.14.

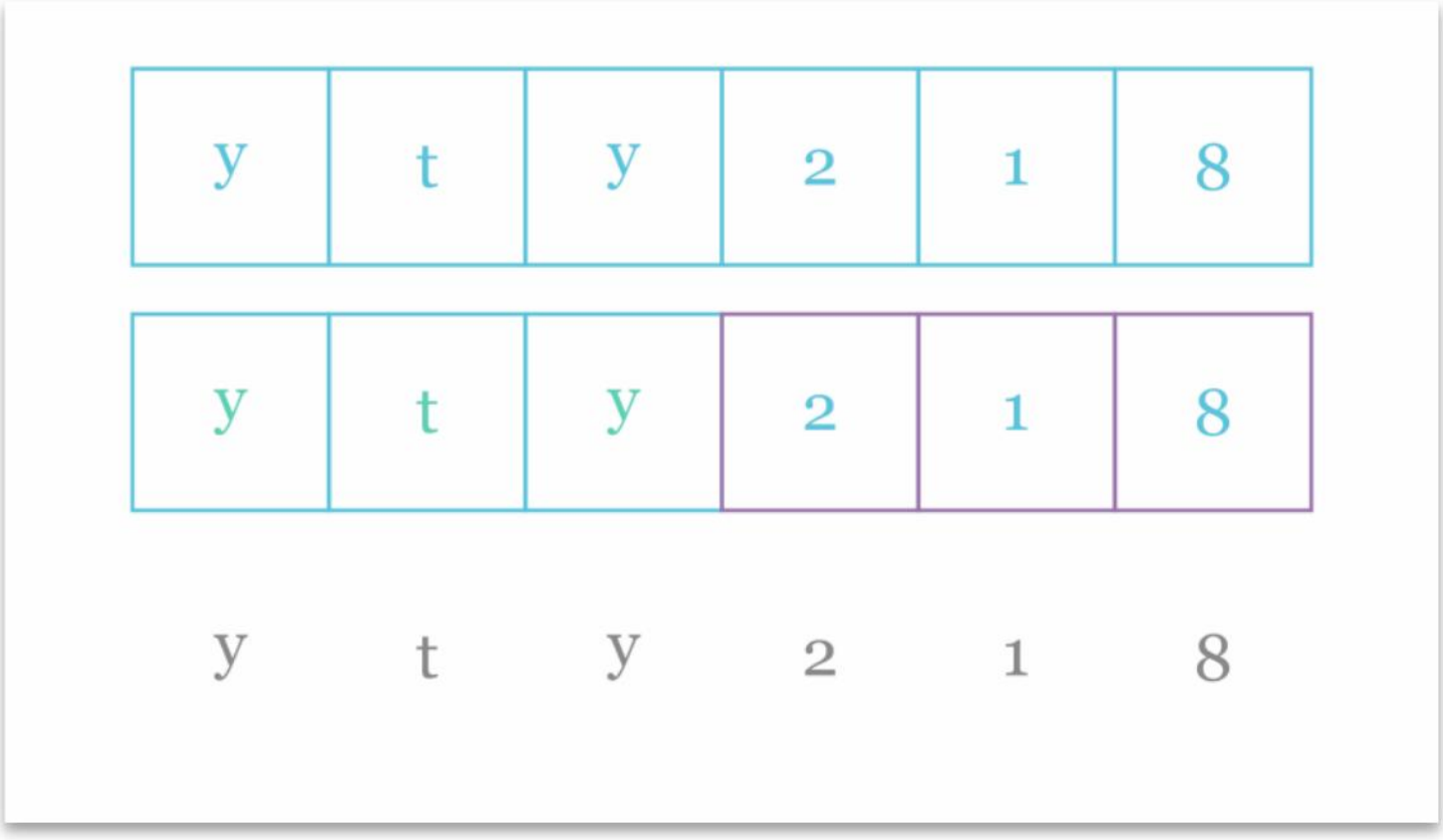

Figure 4.14 SquTex Example diagram

### 4.2.3 Build basic formula animation class

To meet the dynamic visualization needs for determinant calculations, a class called MatrixDet has been constructed. This class inherits from the base matrix structure class MatrixCal and further extends with specialized logic for visual demonstrations. Not only does it support precise extraction of element paths along the main diagonal and secondary diagonal, but it also automatically calculates and renders the intermediate values of each multiplication path, visually highlighting each element involved in the computation one by one. Through color coding, path animations, and numerical annotations, it dynamically presents the process of expanding determinants. The system can automatically generate corresponding calculation expressions (such as a11 × a22 × a33 +...), merge positive and negative terms visually, and ultimately output clear and complete dynamic expansion animations.

Code 2 MatrixCal Constructor

```
Input  : matrix: 2D array, buff: padding, brackets_pair: pair of
         brackets
Output: Constructed matrix with optional brackets
if matrix is not a 2D array then
    Error: Matrix must be a 2D array.;
end
self.matrix ← matrix;
self.buff ← buff;
if brackets_pair is None then
    self.brackets_pair ← ['[', ']']
end
else
    self.brackets_pair ← brackets_pair
end
Call parent constructor;
Call _construct_matrix;
Call _add_brackets with self.brackets_pair;
Function _construct_matrix:
for each row in matrix do
    Create SquTex object with current row;
    Add SquTex object to current object;
end
Arrange objects vertically with zero spacing;
```

This class extracts the set of elements involved in a multiplication step through the get_process_inform() method and records their product value accordingly; it then calls cal_progress_times() to generate a visual expression with an equal sign and a multiplication symbol. The final cal_result_addition() function integrates all process results, dynamically generating addition and subtraction expressions along with the final value. This class works in conjunction with a unified timeline animation control logic to fully reproduce the entire process from element selection to calculation results of determinants. The specific effect is shown in Figure 4.15.

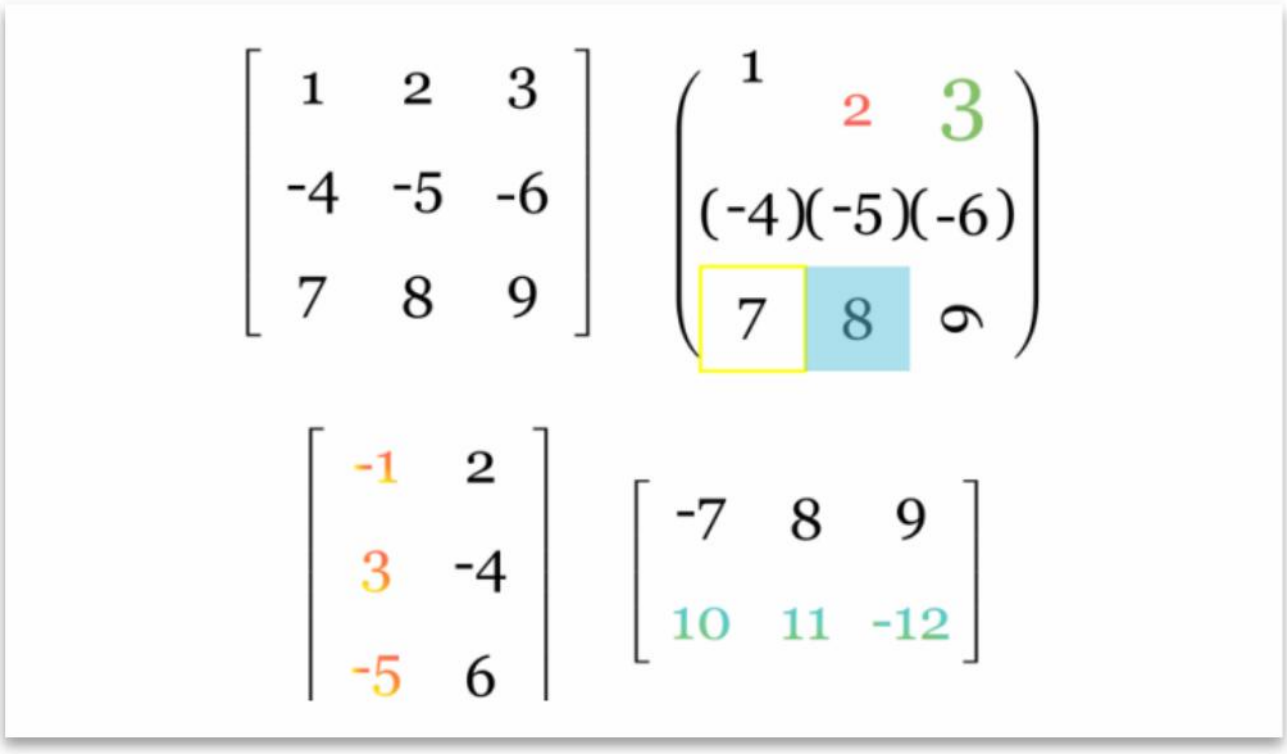

Figure 4.15 MatrixCal Example diagram

### 4.2.4 Monocular operation is divided into step-by-step animation construction

In view of the dynamic visualization requirements for determinant calculation, a class MatrixDet is constructed, which inherits from MatrixCal and adds the functions of extracting the product path of the main diagonal and the secondary diagonal and visualizing the intermediate process.

Code 3 MatrixDet Constructor

```
Input  : matrix: 2D array
Output: Constructed object for matrix determinant calculation
Call parent constructor with matrix and brackets_pair = ['—', '—'];
if matrix is not square then
    Error: Matrix is not square, cannot compute determinant.;
    Raise ValueError with error message;
end
self.res ← 0;
self.res_lst ← empty list;
```

This class extracts the set of elements involved in a multiplication step through the get_process_inform() method and records their product value accordingly; it then calls cal_progress_times() to generate a visual expression with an equal sign and a multiplication symbol. The final cal_result_addition() function integrates all process results, dynamically generating addition and subtraction expressions along with the final value. This class works in conjunction with a unified timeline animation control logic to fully reproduce the entire process from element selection to calculation results of determinants. The specific effect is shown in Figure 3.16.

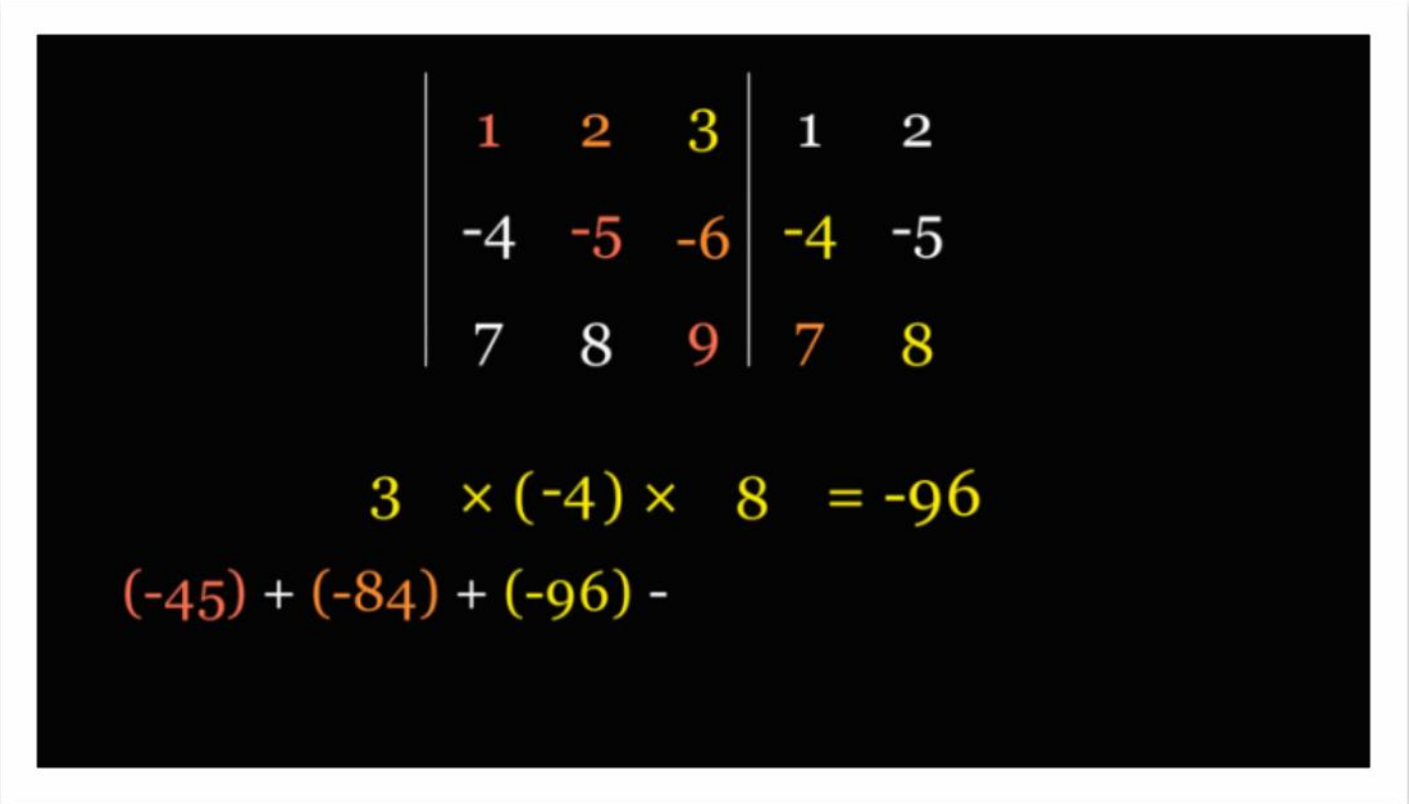

Figure 4.16 An example of one frame of MatrixDet

Classes are instantiated an MatrixDetShowSceneMatrixDet d dynamically displayed through inheritance. The algorithm realizes the step-by-step selection of diagonal columns for cumulative multiplication, then records the results step by step, and finally displays the results. The algorithm flowchart of the specific implementation is shown in Figure 4.17, and the class method call relationship diagram is shown in Figure 4.18.

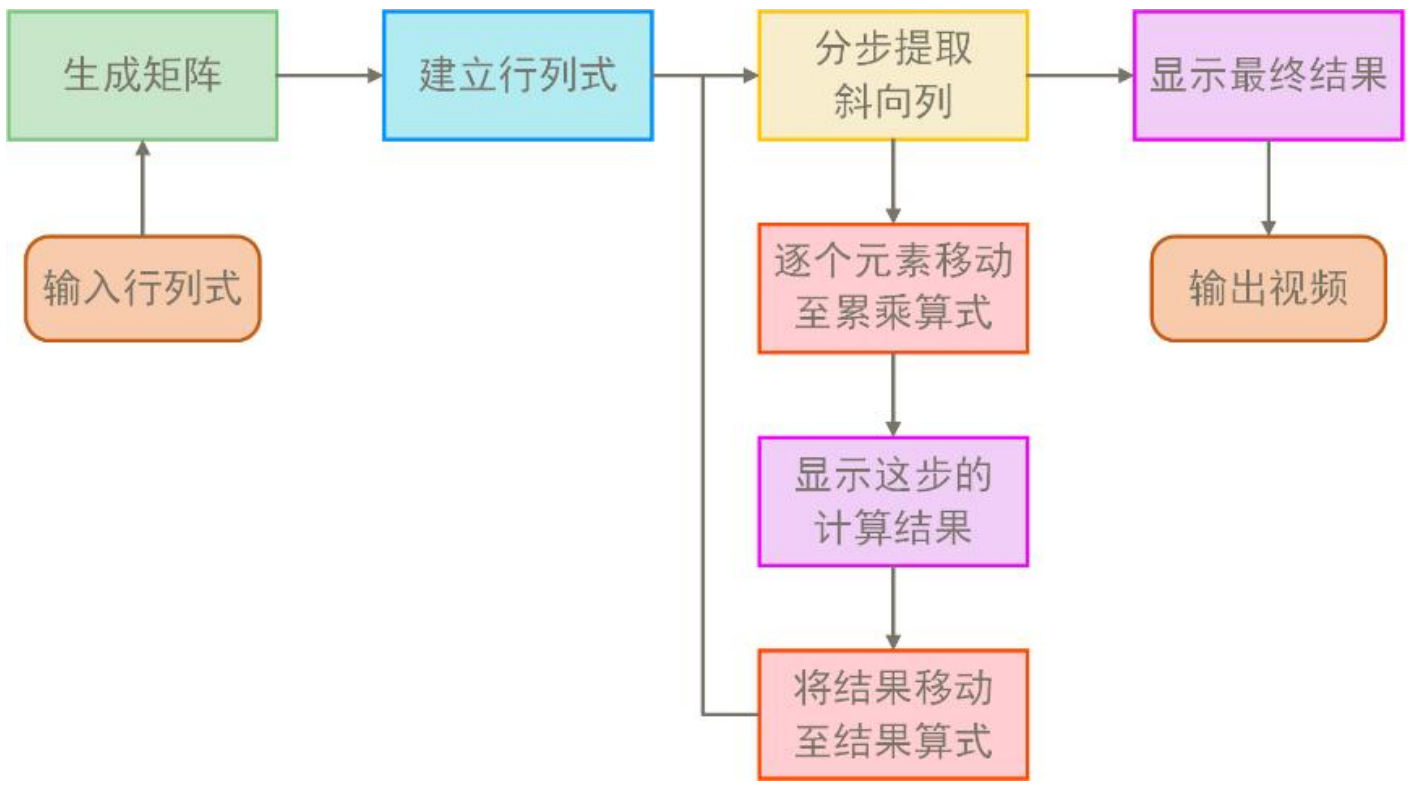

Figure 4.17 Step-by-step algorithm flow chart for determinant

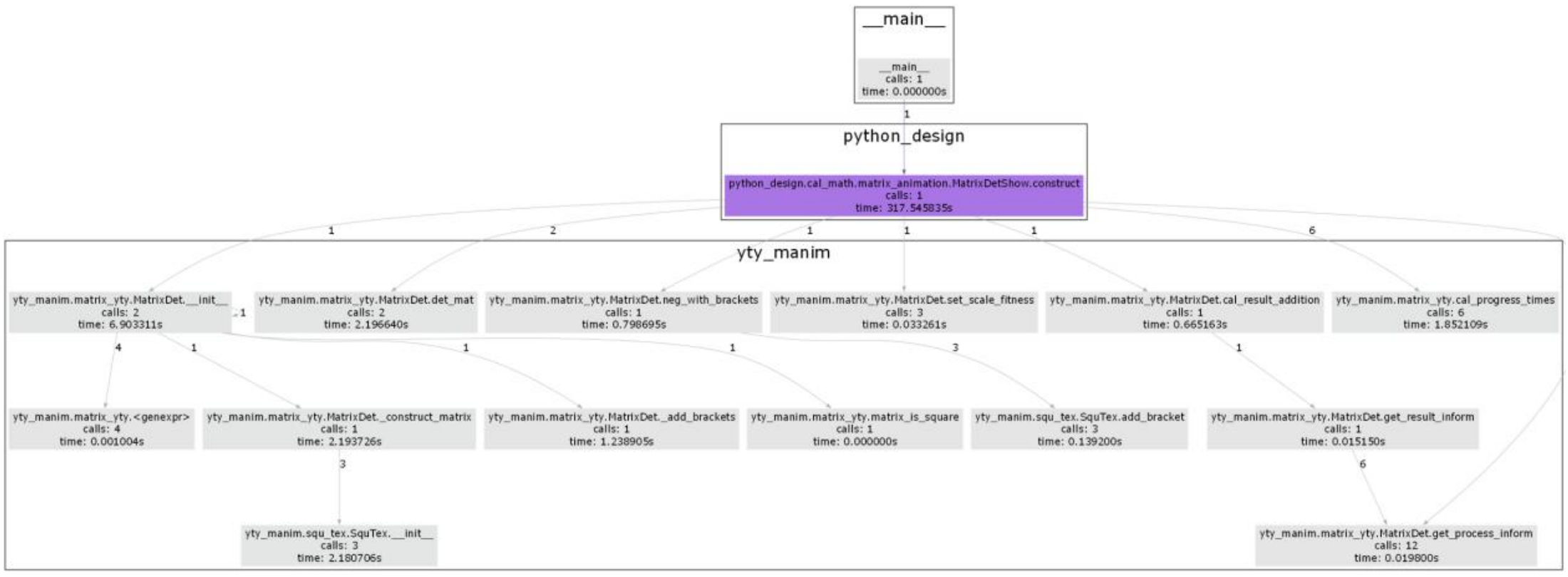

Figure 4.18 Step-by-step demonstration of the method call relationship for the determinant

## 4.2.5 Two-eye operation is divided into step-by-step animation construction

In order to support the step-by-step animation demonstration of matrix addition and multiplication, the system designs a class called MatrixMath, which extends from MatrixCal and introduces two matrices as operation objects. The specific construction function is shown in code 5.

Code 5 MatrixMath Constructor

**Input** : matrix: 2D array, forward_mat: optional 2D array, backward_mat: optional 2D array, kwargs: additional arguments

**Output:** Constructed object with optional forward and backward matrices

```
Call parent constructor with matrix and kwargs;
self.forward_mat ← forward_mat;
self.backward_mat ← backward_mat;
```

This class internally encapsulates the `addition_mat()` and `dot_multiplication_mat()` methods, which perform the element-wise addition and matrix multiplication logic of two matrices, respectively, and output the corresponding visualized matrices. Through the `get_mul_progress()` function, the system generates intermediate steps in expression form for each computational unit, demonstrating how the products of corresponding row and column elements are added element-wise to form individual elements of the result matrix, thereby achieving an animated recreation of the multiplication process. During both addition and multiplication, this class employs operations such as matrix block selection, element highlighting, and result movement, enabling users to gain a direct understanding and deeper insight into the mechanism of matrix operations. The specific addition effect is shown in Figure 4.19, and the multiplication effect is shown in Figure 4.20.

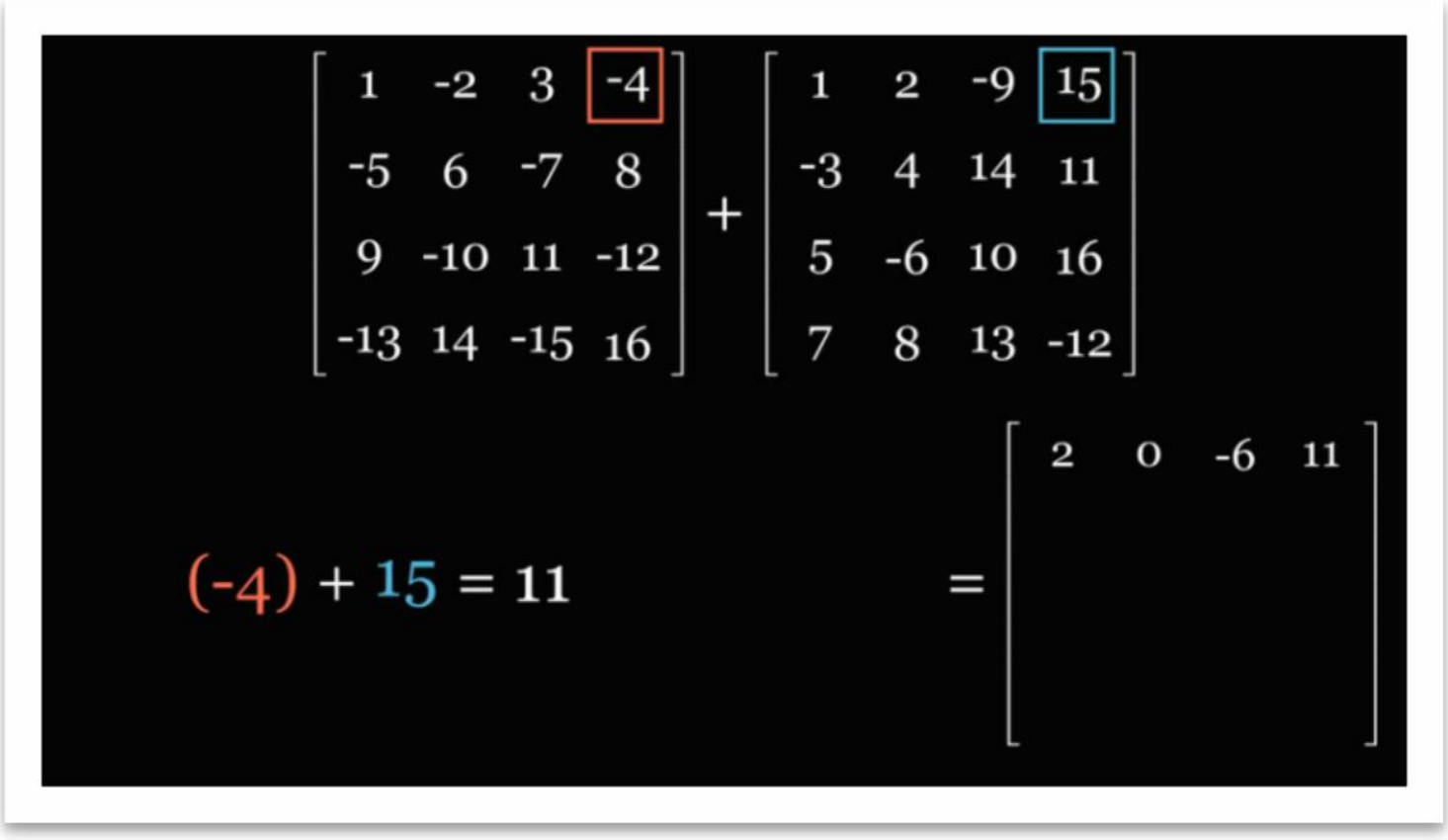

Figure 4.19 An example of a frame of addition in MatrixMath

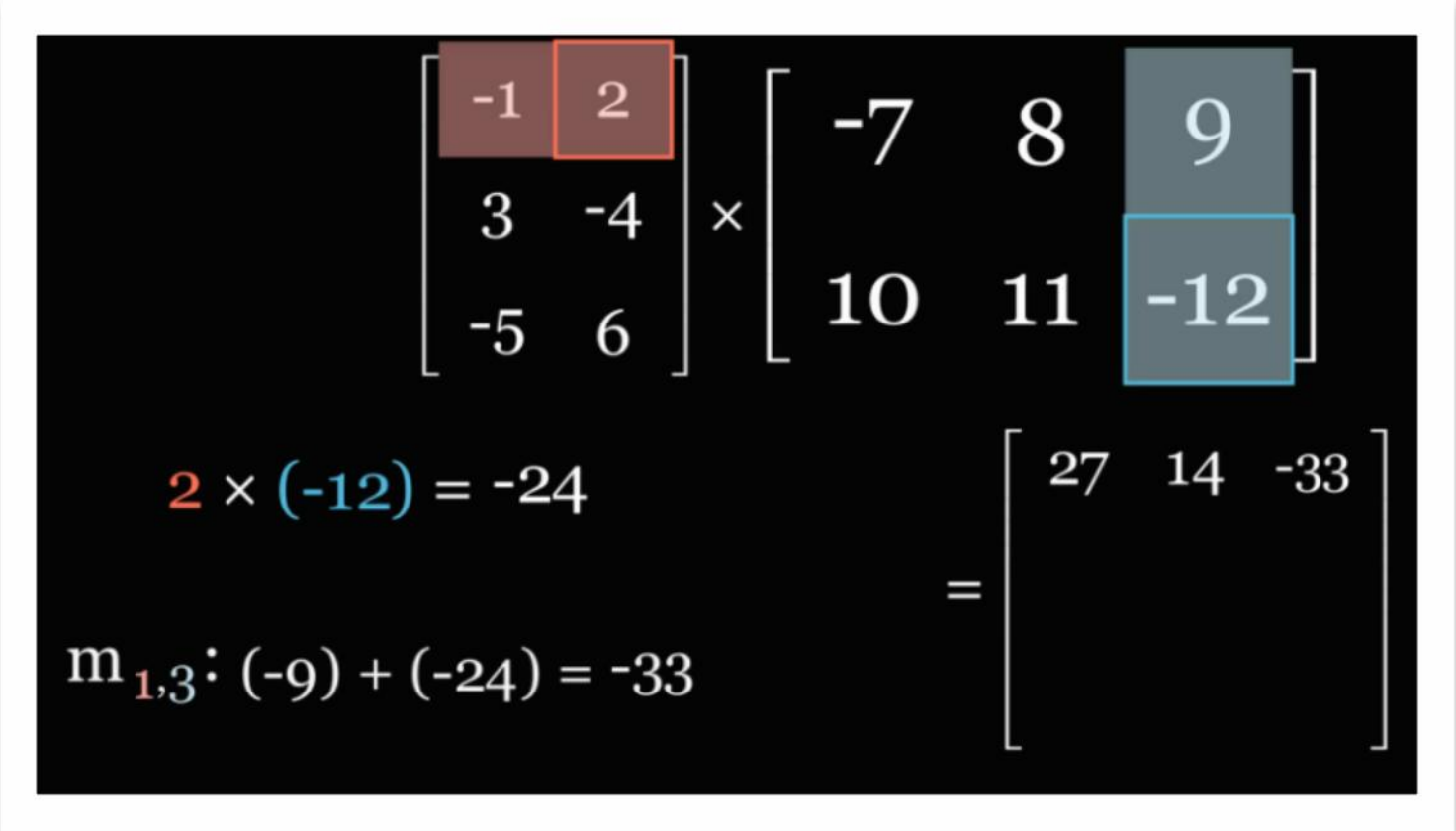

Figure 4.20 An example of a frame of multiplication by MatrixMath

Classes are implemented through inheritance, MatrixAddtionShowSceneMatrixMathaddition_mat where methods in the class are instantiated and dynamically displayed. This achieves the element selection of the original dual matrix in steps, and after addition, the result is moved to the result matrix. The specific algorithm flowchart is shown in Figure 4.21, and the class method call relationship diagram is shown in Figure 4.22.

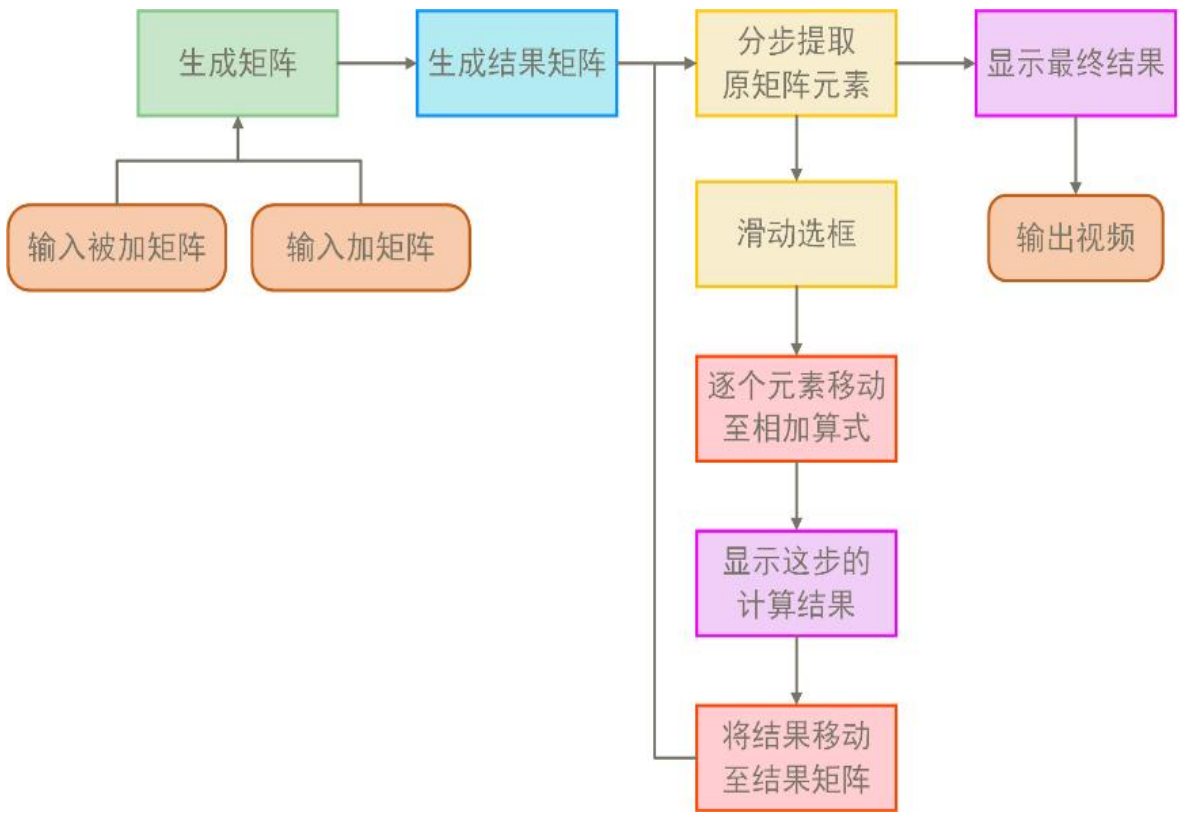

Figure 4.21 Flowchart of the algorithm for step-by-step demonstration of matrix addition

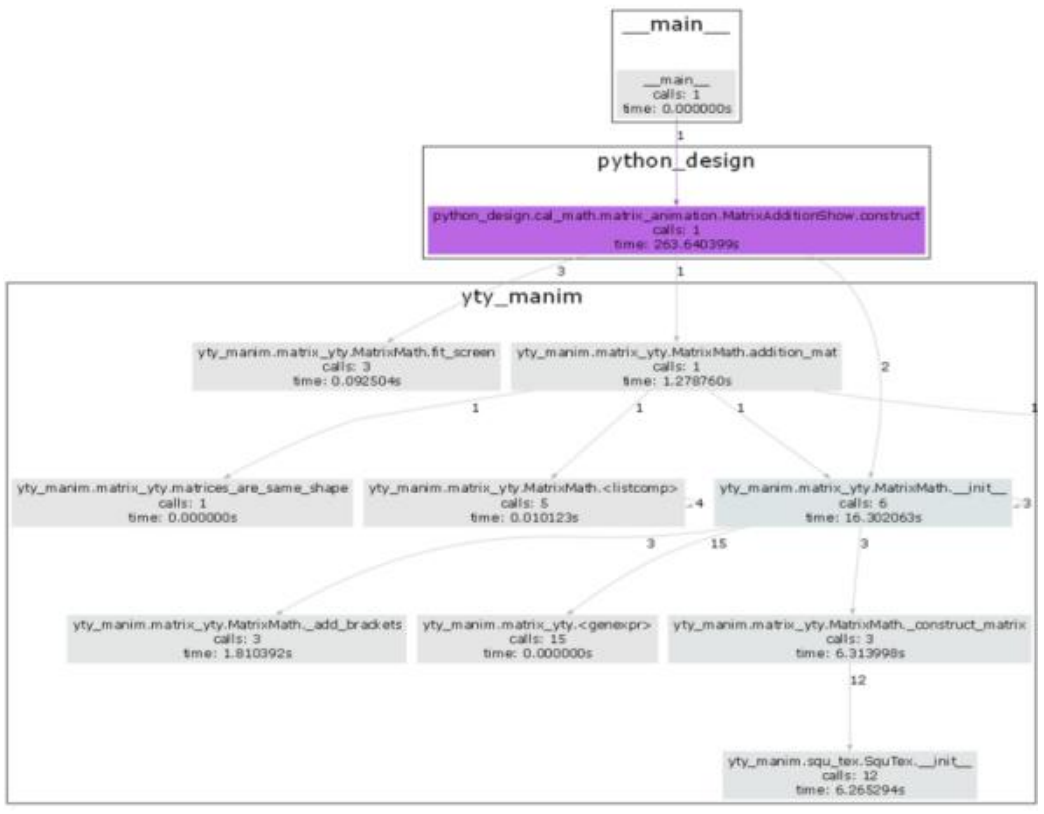

Figure 4.22 Step-by-step demonstration of matrix addition class method call relationships

Inheritance and instantiation of methods in the class achieve dyMatrixMulShowSceneMatrixMathdot_multiplication_matm$_{ij}$i, jj, iijnamic display, implementing step-by-step operations during computation. The first element of the first matrix is selected and multiplied by the corresponding element of the second matrix, with the result moved to the addition formula. After completing the calculations for the respective rows and columns of the two matrices, the sum is moved to the result matrix. The specific algorithm flowchart is shown in Figure 4.23, and the class method call relationship diagram is shown in Figure 4.24.

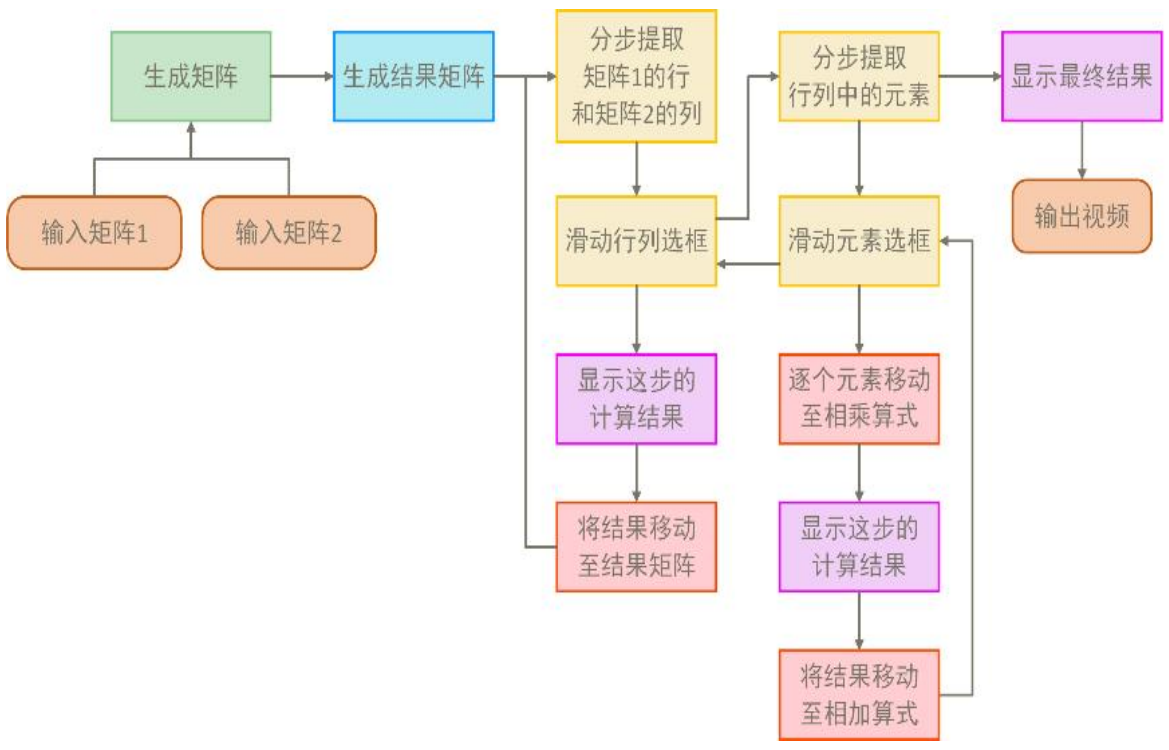

Figure 4.23 Flowchart of the algorithm for matrix multiplication step by step

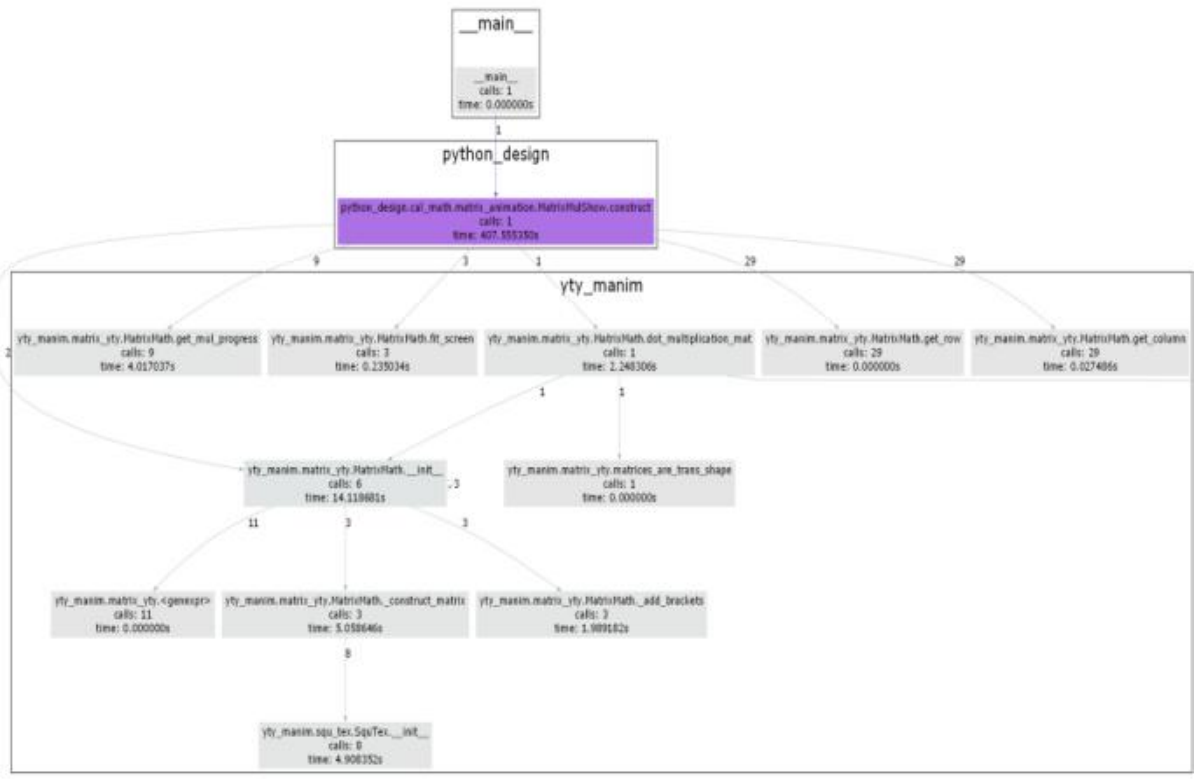

Figure 4.24 Step-by-step demonstration of matrix multiplication class method call relationship

# 5 experiment

## 5.1data

The original data set "Handwritten math symbols dataset" has 82 classes, including their own operation symbols and numbers and letters. Each class contains about 10,00045x45 jpg black and white images.

The handwritten matrix data set is generated by the algorithm in Chapter 3.2 to form a YOLO format data set.

## 5.2benchmark model

YOLO is a single-stage object detection model proposed by Joseph Redmon in 2015. The YOLO algorithm redefines the object detection task, treating it as a single regression problem, directly mapping from image pixels to bounding box coordinates and class probabilities. In the YOLO model, the input image is divided into equally sized grids. For each grid, if the center of an object falls within it, the detection and prediction of that object are handled by that grid, ultimately outputting the bounding box and the corresponding class probability. The innovation of YOLO lies in its ability to predict and classify the entire image without having to assign an object category or background to each region as other models do. This reduces the likelihood of background mispredictions and enhances the detection capability for small objects.

The network architecture of YOLOv11 is divided into Backbone, Neck and Head. Compared with the previous model of YOLO, the C3K2 module and C2PSA module are improved on the basis of C2F. Figure 2.1 shows the network architecture of YOLOv11.

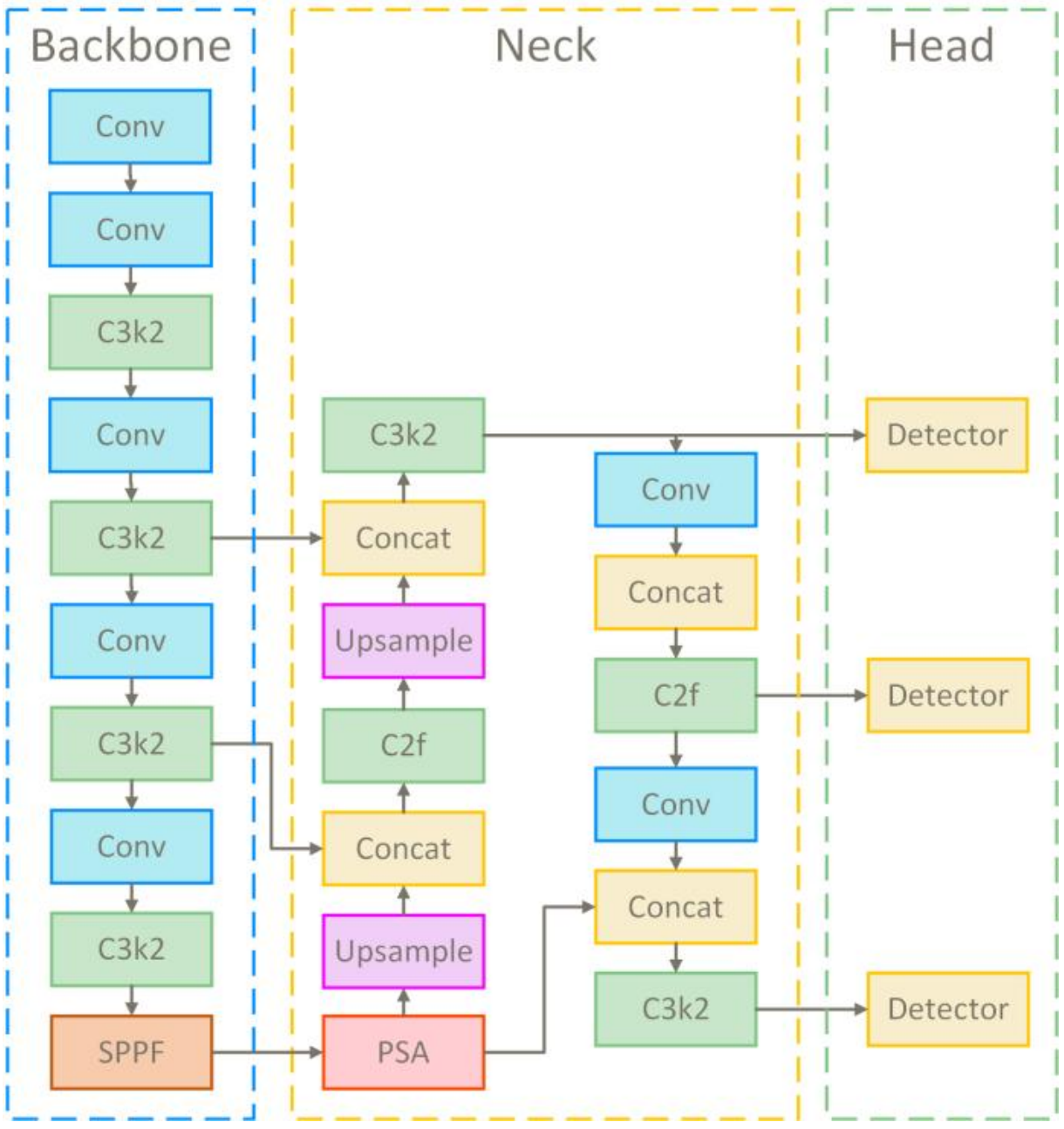

Figure 5.1 YOLOv11 architecture diagram

The network extracts basic features of images, typically capturing spatial and semantic information from low to high levels through convolutional neural networks. The Neck network performs multi-scale feature fusion on the feature pyramid, ensuring that the model can capture both small and large targets simultaneously. This enhances the expressive power of features and bridges the gap between the Backbone and Head modules. The Head network outputs the final target category and bounding box coordinates, performing localization and classification of targets using a specially designed anchor method. The core structure of the C3K2 module is CSPNet (Cross Stage Partial Network). The C3K2 module introduces multi-scale convolutional kernels C3K, where K represents an adjustable kernel size, such as 3x3, 5x5, etc. This design expands receptive fields, enabling the model to capture broader contextual information. In Backbone, C3K2 is used for efficient extraction of basic features, capturing both low-level and high-level semantic features. In Neck, the feature pyramid structure performs feature fusion through C3K2, ensuring detection capability for small targets and localization across scales. The core of the C2PSA module is PSABlock, a module with self-attention mechanisms, also known as the transformer structure. Adding this module enhances the ability of Backbone to extract features. The SPPF module achieves fixed-size output by performing different sizes of pooling operations on feature maps and integrating the results. This technique can effectively deal with targets of different sizes, and improve the generalization ability and robustness of neural network.

## 5.3appraisal procedure

### 5.3.1 Mean average accuracy

Average Precision (mAP, mean Average Precision): mAP is the most commonly used evaluation metric in object detection. It calculates the average precision (AP) for each category and averages it across all categories. In the evaluation of YOLO algorithms, mAP is typically calculated at different IoU thresholds. IoU (Intersection over Union) is the ratio of the overlapping area between a predicted box and a ground truth box, with 0.5 often used as the standard threshold for evaluation. However, some studies also use an IoU threshold of 0.75 or higher.

AP Calculation: For each category, calculate the precision and recall of predicted boxes at different confidence (confidence) thresholds. Precision is the ratio of correctly predicted boxes to all predicted boxes, and recall is the ratio of correctly predicted boxes to all true boxes. AP is the area under the precision-recall curve, representing the model's overall detection capability across different thresholds.

### 5.3.2 accuracy

Precision is the ratio of the target box correctly predicted to all predicted boxes. High precision means that most of the predicted boxes are correct targets and fewer false positives occur.

$$Precision = \frac{TP}{TP + FP} \tag{7}$$

Among them, TP is the number of true positives and FP is the number of false positives.

### 5.3.3 recall

The recall rate is the ratio of the correct predicted boxes to all the real ones. A high recall rate means that the model can recognize most of the targets.

$$Recall = \frac{TP}{TP + FN} \tag{8}$$

Among them, TP is the number of true positives and FN is the number of false negatives.

### 5.3.4 F1 value

The F1 value is the harmonic average of precision and recall, which comprehensively evaluates the performance of a model in terms of precision and recall. Models with high F1 values are usually a good balance between precision and recall.

$$F1 = 2 \times \frac{Precision \times Recall}{Precision + Recall} \tag{9}$$

## 5.4Experimental configuration

### 5.4.1 experimental platform

The experiment was conducted on a hardware platform equipped with NVIDIA RTX 4090 GPU, utilizing the distributed computing resources provided by AutoDL servers. RTX 4090 offers robust computational performance, significantly accelerating the training process of the YOLO model. The experimental platform includes AMD Ryzen 95900X CPU, 64GB DDR4 of memory, 2TB NVMe SSD of storage, and runs the Ubuntu 20.04 operating system, equipped with PyTorch 2.0 and CUDA 11.3, ensuring efficient execution of deep learning tasks.

### 5.4.2 Hyperparameter Settings

The input size used is 640x640, with the AdamW optimizer. The initial learning rate is set to 0.001, and a CosineAnnealing learning rate scheduler is used to gradually reduce the learning rate. The batch size for training is 16, which has been tested as the optimal configuration on RTX 4090 GPU. The loss function includes classification loss, localization loss, and confidence loss, ensuring the accuracy of the model in detection tasks. During training, an IoU threshold of 0.5 is used to determine if a detection is correct.

### 5.4.3 Training process

During training, each epoch model is validated and mAP, precision, recall and other indicators are calculated. During training, data enhancement techniques such as random cropping, scaling and flipping are used to improve the generalization ability of the model.

# 5.5experimental result

YMCAnet excels in handwritten equation recognition tasks, with specific SOTA data shown in Table 5.1. Its mAP@5 reaches 94.5%, and mAP@95 reaches 88.2%, significantly outperforming other mainstream models such as YOLOv5 (91.0% and 82.0%) and Mask R-CNN (90.1% and 82.5%). This indicates that YMCAnet has a stronger advantage in high-precision recognition tasks, especially in detail handling and complex scenarios. Additionally, YMCAnet's parameter size is 35.2 million, smaller than YOLOv5 (46.5 million) and Mask R-CNN (47.1 million), making it more efficient in computational resources, suitable for resource-constrained devices. Meanwhile, YMCAnet's computational complexity is 120.5 GFLOPs, lower than YOLOv5 (125.8 GFLOPs) and Mask R-CNN (235.3 GFLOPs), ensuring efficient inference speed and real-time processing capabilities. In contrast, although CRNN and Tesseract OCR perform well in terms of parameters and computational complexity, they do not match YMCAnet's performance in mAP, particularly in high-precision handwritten equation recognition tasks, where they cannot provide the same level of accuracy. Therefore, YMCAnet shows great advantages in balancing accuracy, calculation efficiency and parameter quantity, especially suitable for high requirements of handwritten formula recognition scenarios.

Table 5.1 YMCAnet SOTA performance table of other related models

| Model | Input Size | mAP@.5 | mAP@.9 | Params(M) | GFLOP |
|---|---|---|---|---|---|
| YOLOv5[1] | 640x640 | 91.0 | 82.0 | 46.5 | 125.8 |
| EAST[8] | 640x640 | 85.5 | 77.5 | 36.3 | 110.7 |
| CRNN[9] | 32x128 | 90.2 | 80.5 | **17.8** | **48.3** |
| ResNet + LSTM[10] | 128x32 | 88.5 | 75.6 | 25.6 | 61.2 |
| Attention OCR[11] | 256x256 | 87.1 | 78.4 | 48.9 | 131.2 |
| Faster R-CNN[12] | 800x800 | 89.4 | 81.0 | 39.4 | 205.6 |
| Mask R-CNN[13] | 800x800 | 90.1 | 82.5 | 47.1 | 235.3 |
| VGG16 + RNN[14] | 224x224 | 83.2 | 74.1 | 138.4 | 142.7 |
| Tesseract OCR[15] | Variable | 85.0 | 77.8 | 16.2 | 30.4 |
| SAST[16] | 640x640 | 86.7 | 78.3 | 40.0 | 118.0 |
| **DeepText[17]** | 32x128 | 87.3 | 80.0 | 21.9 | 50.0 |
| TextBoxes++[18] | 640x640 | 88.0 | 80.5 | 38.1 | 112.3 |
| **YMCAnet(MINE** | 640x640 | **96.5** | **89.2** | 35.2 | 120.5 |

YMCAnet demonstrated the best performance, maintaining an mAP@ above

94% in most categories. In particular, categories 2, 9, and 0 achieved high precision rates of 97.5%,97.2%, and 98.0%, respectively. By comparison, YMCAnet outperformed other models in most numerical and symbolic categories, especially in capturing complex edges, handwriting, and symbols, significantly outperforming YOLOv5, EAST, CRNN, and Tesseract OCR. Its accuracy is not only evident in common numerical categories but also stands out in symbol recognition (such as +, -, /), giving YMCAnet a significant advantage when handling handwritten equation recognition tasks.

Table 5.2 mAP@ of some specific categories of YMCAnet and other related models.5

Comparison

| model | 1 | 2 | 3 | 4 | 5 | 6 | 7 | 8 |
|---|---|---|---|---|---|---|---|---|
| **YOLOv5** | 94.2 | 96.3 | 94 | 94.5 | 93.8 | 94.9 | 95.5 | 94.7 |
| **EAST** | 93.1 | 94 | 92.5 | 93.2 | 92.4 | 93.5 | 94.1 | 93.3 |
| **CRNN** | 90.4 | 91.7 | 90.3 | 91.1 | 90 | 91.9 | 91.6 | 91 |
| **Tesseract OCR** | 87.5 | 88 | 86.5 | 87 | 86.2 | 87.3 | 87.8 | 86.7 |
| **YMCAnet(MINE)** | **96.1** | **97.5** | **95.8** | **96.3** | **95.2** | **96.7** | **97.2** | **96.4** |

| model | 9 | 0 | + | - | * | / | [ | ] |
|---|---|---|---|---|---|---|---|---|
| **YOLOv5** | 95.8 | 96.9 | 92.8 | 93.2 | 92.5 | 93.4 | 91.3 | 91.4 |
| **EAST** | 94.4 | 94.9 | 91.4 | 91.7 | 90.8 | 92.1 | 90.3 | 90.5 |
| **CRNN** | 92.2 | 92.5 | 88.9 | 89.5 | 88.4 | 89.6 | 87.7 | 88 |
| **Tesseract OCR** | 88.4 | 89.2 | 84.5 | 85 | 84.1 | 85.2 | 83.3 | 84 |
| **YMCAnet(MINE)** | **97.2** | **98.0** | **94.1** | **94.4** | **93.7** | **94.8** | **92.9** | **93** |

# 6 Analysis and discussion

Table 6.1 Ablation status of YMCAnet

| model | mAP@.50 | mAP@.95 |
|---|---|---|
| **Baseline** | 92.4 | 84.3 |
| **Baseline + Mamba** | 94.6 | 86.7 |
| **Baseline + CA** | 95.1 | 87.3 |
| **YMCAnet(MINE)** | **96.5** | **89.2** |

As the benchmark model, baseline network does not introduce any enhancement modules. Its accuracy is relatively average, especially in recognizing complex edges and handwritten symbols. The mAP@.50 value is high, indicating that the model can recognize most characters at lower IoU thresholds. However, when the threshold is raised to mAP@.95, the performance of the model significantly drops, particularly at higher IoU thresholds, where it fails to effectively identify small targets or characters with blurred edges, suggesting that the base model has weak capabilities in capturing fine details.

Baseline + Mamba builds upon the baseline model by incorporating a Mamba backbone network, which focuses on enhancing the overall feature representation capability of the model. The Mamba network improves the ability to capture complex textures and details through deeper convolutional layers and feature extraction. Results show that performance has significantly improved at mAP@.50, increasing from 92.4% to 94.6%. Particularly in simpler number and symbol recognition, the model demonstrates stronger recognition capabilities. At mAP@.95, the performance improvement is also notable, reaching 86.7%, an increase of about 2.4% compared to baseline. This indicates that the introduction of the Mamba backbone network significantly enhances the model's robustness at higher IoU thresholds, especially for small targets and edge-blurred characters.

After introducing the CoordAttention module, the spatial localization capability of the model has significantly improved. The CA module enhances attention mechanisms by incorporating coordinate information, enabling the model to more accurately locate characters in the image, especially when handling complex handwritten symbols (such as +, -, /). The mAP@.50 improved to 95.1%, a 2.7% increase compared to baseline. This improvement highlights the significant role of the coordinate attention mechanism in enhancing overall recognition accuracy. On the mAP@.95, the CA module also delivered notable performance gains, reaching 87.3%, a 3% increase over baseline. This further validates the effectiveness of the CA module, particularly under stricter precision requirements, where it can more precisely identify details and small targets.

YMCAnet (Mamba + CA) combines Mamba, the backbone network, and the CoordAttention module, making it the most powerful model among the four configurations. By integrating these two advanced modules simultaneously, YMCAnet achieves an mAP@50 of 96.5%, a 4.1% improvement over baseline. This significant boost indicates that combining the Mamba network with the CA module enhances the model's ability to capture characters, symbols, and complex edges comprehensively, especially in more challenging handwriting formula and symbol recognition tasks, where its performance is notably superior. At mAP@95, YMCAnet performs even better, reaching 89.2%, a 4.9% improvement over baseline, demonstrating its strong capability in high-precision tasks. Compared to adding Mamba or CA modules individually, YMCAnet maintains stable accuracy at high IoU thresholds, particularly excelling in handling small targets and blurry symbols.

# 7 Interface for interaction

The visual interface streamlitmanimwritten by the package is used to realize the connection of YOLO detection, animation, user IO operation and file storage to realize a complete function. The specific operation flow chart is shown in Figure 7.1, and the generated matrix file storage structure is shown in Figure 7.2. For details, see the website:

https://animatecal-aesrxwe852bslylhgvrfxx.streamlit.app/。

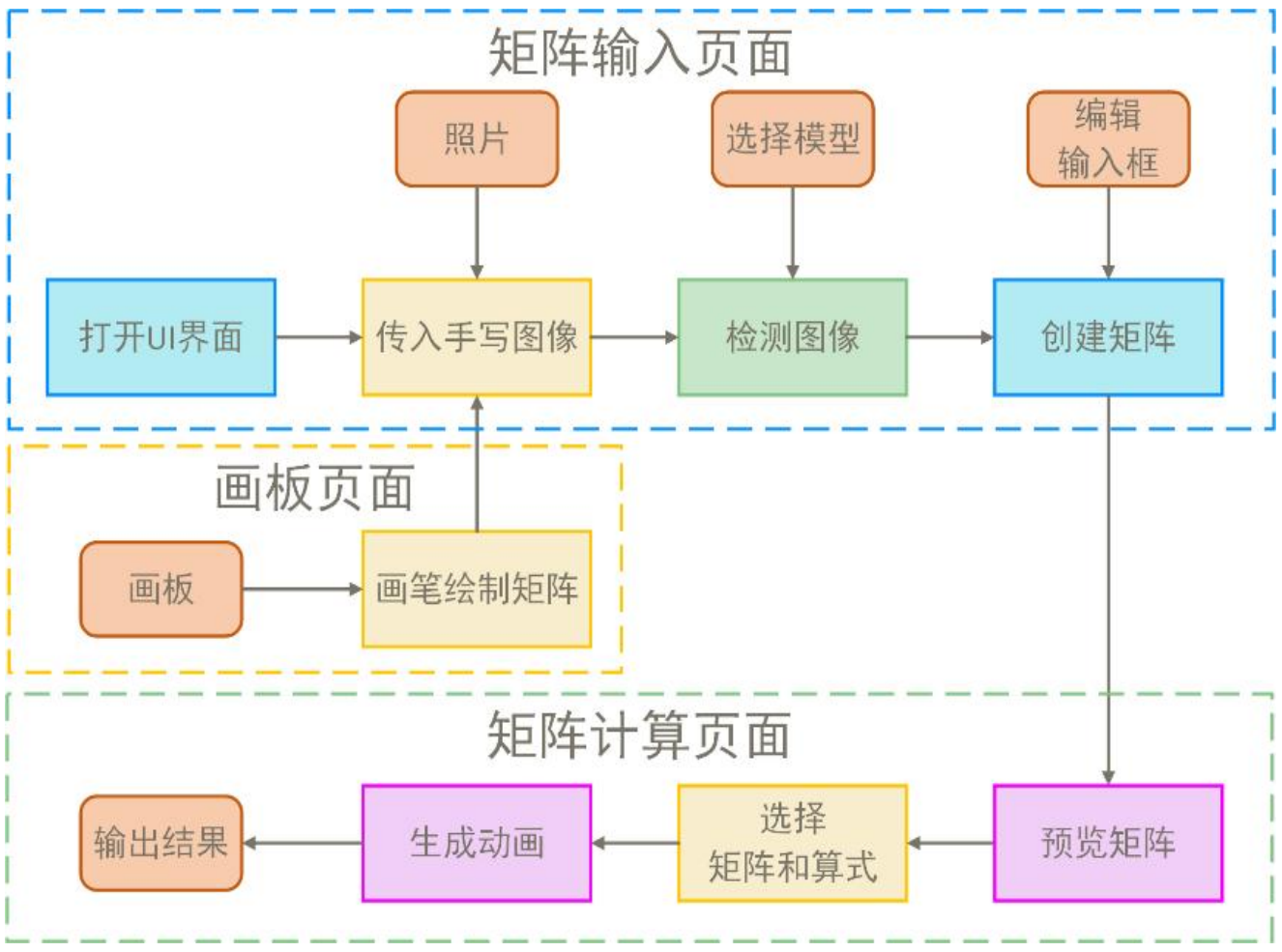

Figure 7.1 User operation flow chart

```
saves
├── manim
│   └── manim_cache.png
└── math_cache
    ├── cache.txt
    ├── matrix0_cache.txt
    ├── matrix0_cache.png
    ├── matrix1_cache.txt
    ├── matrix1_cache.png
    ├── canvas_matrix.txt
    ├── canvas_cache.png
    ├── canvas_cache.ps
    └── ...(其他矩阵存储)
```

Figure 7.2 Generation of matrix file storage structure

## 7.1Mathematical formula input interface

The input page of mathematical formula contains six functions, as shown in Table 1. It deals with the process of matrix recognition and input.

Table 1 Function table of formula input page

| Function table of the formula input page | |
|---|---|
| The original image shows | Display images, and users can select images in the system through the matrix input control panel |
| Detect images | Click the "Detect Image" button on the control panel to display the matrix after clustering segmentation |
| Edit the formula | Display the recognition results, and the user can modify the recognized matrix |
| Create the formula | After clicking on "Create Matrix" in the control panel, you need to name the matrix. The window will display the latex format of the matrix after creation |
| preference pattern | Select the model version through the drop-down box. If an image is selected, the new selected model will be used to automatically detect the image after selection |
| skip | Jump to other pages and draw matrices in the drawing board |

Using to build the framework, FrameButtonmanimosnumpy classes make it easy to add a button that binds to a function, facilitating access to external functions and class object interfaces I previously reserved, such as YOLO detection and animations. Additionally, libraries are used for some IO operations. Here, caching is employed to store all matrix information from this operation, with specific matrix content stored in cache.txt files using the format. The detailed interface and floating control panel are shown in Figures 7.3 and 7.4.

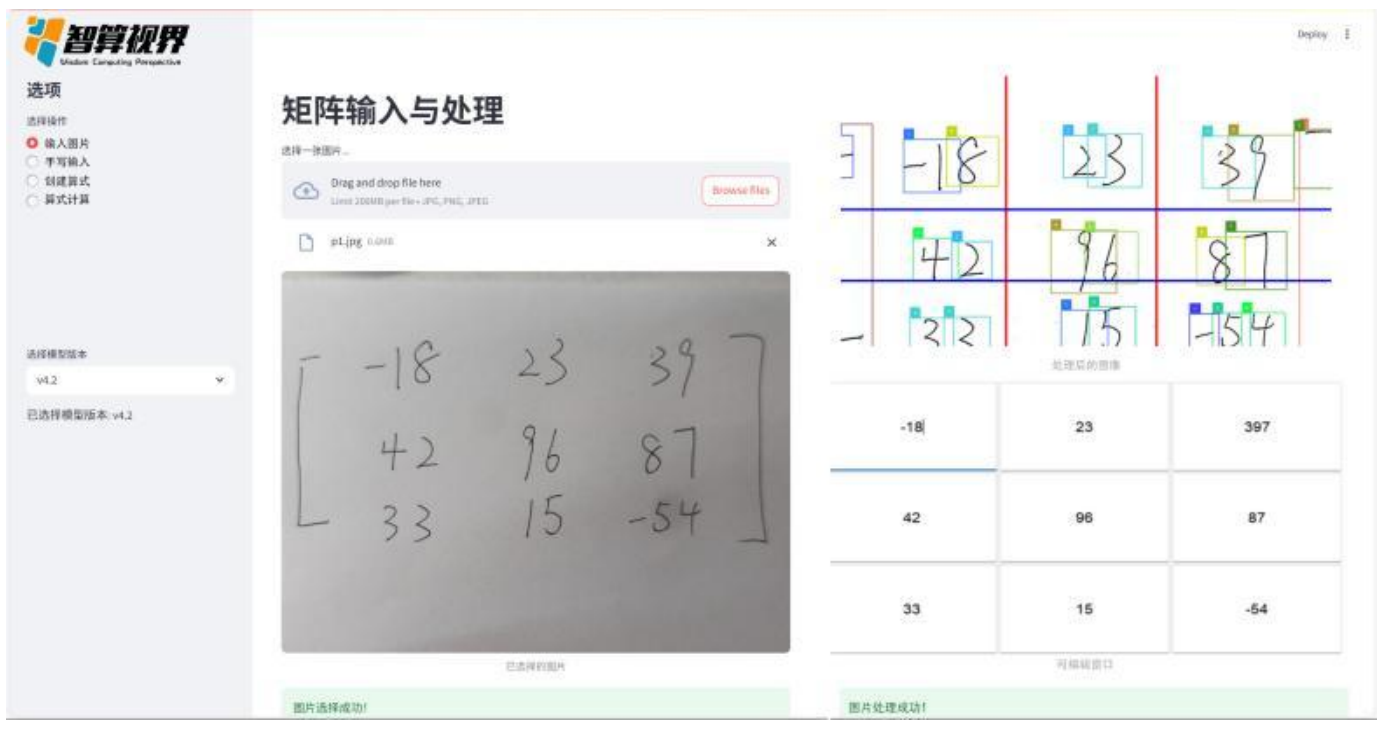

Figure 7.3 Input interface of mathematical formula

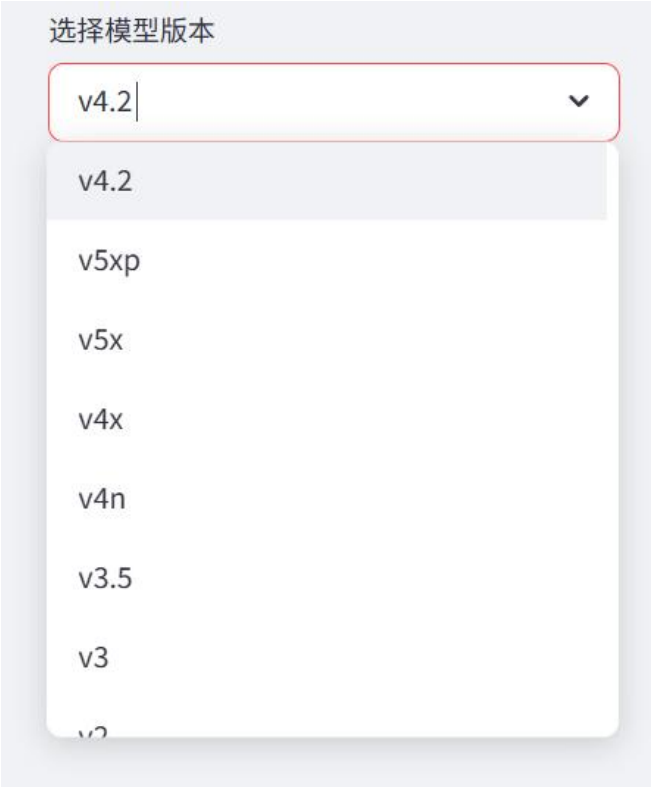

Figure 7.4 Matrix floating control panel

## 7.2Mathematical formula calculation interface

Figure 7.5 is the mathematical calculation page, which contains four functions, as shown in Table 2. After the matrix input is completed, this page is entered to complete the step-by-step visualization process of matrix calculation.

Table 2 Function table of calculation page

| Function table of the calculation page | |
|---|---|
| Preview the formula | Select the generated formula from the list of formulas to preview the selected formula |
| mode selection | Select the calculation mode (determinant, matrix addition, matrix multiplication), and then fill in the selected matrix into the calculation box |
| Generate video | Click the Generate Video button to automatically play a video in the visualization presentation window that contains the calculation process for each element in the matrix |
| result display | Display the calculation results in latex format (the calculation results have been verified by numpy) |

By reading the stored png images to preview the scrollbar, after clicking the lock button, the first matrix is recorded as matrix0_cache.txt, and the second matrix manimcv2VideoCapturenumpy follows the same process. By reading from the cache, videos corresponding to the selected mode are generated. The video is read using the following method, then converted to the appropriate format, displayed frame by frame, and a pause function is added. The specific content of the matrices is cached in cache.txt files using the specified format.

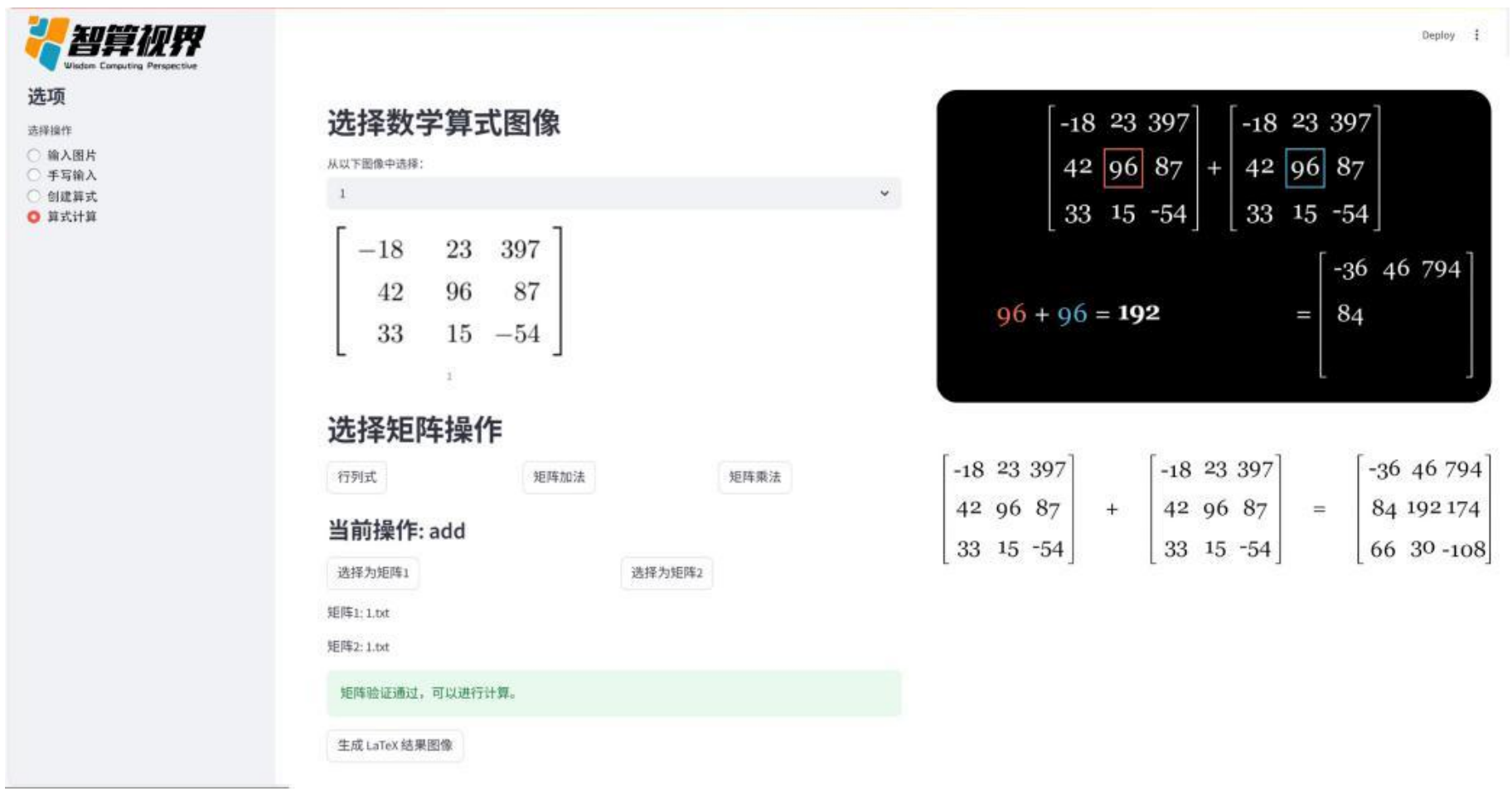

Figure 7.5 Mathematical formula calculation page

## 7.3Free input board interface

The class is called to Canvasdrawmodeimplement the basic canvas construction and brush functions, set the foreground color (black) as the brush, and the background color (white) as the eraser. The brush and eraser share a function, but the size is controlled by a global variable to determine whether it is a brush or an eraser mode. See Figure 7.6 for the specific interface.

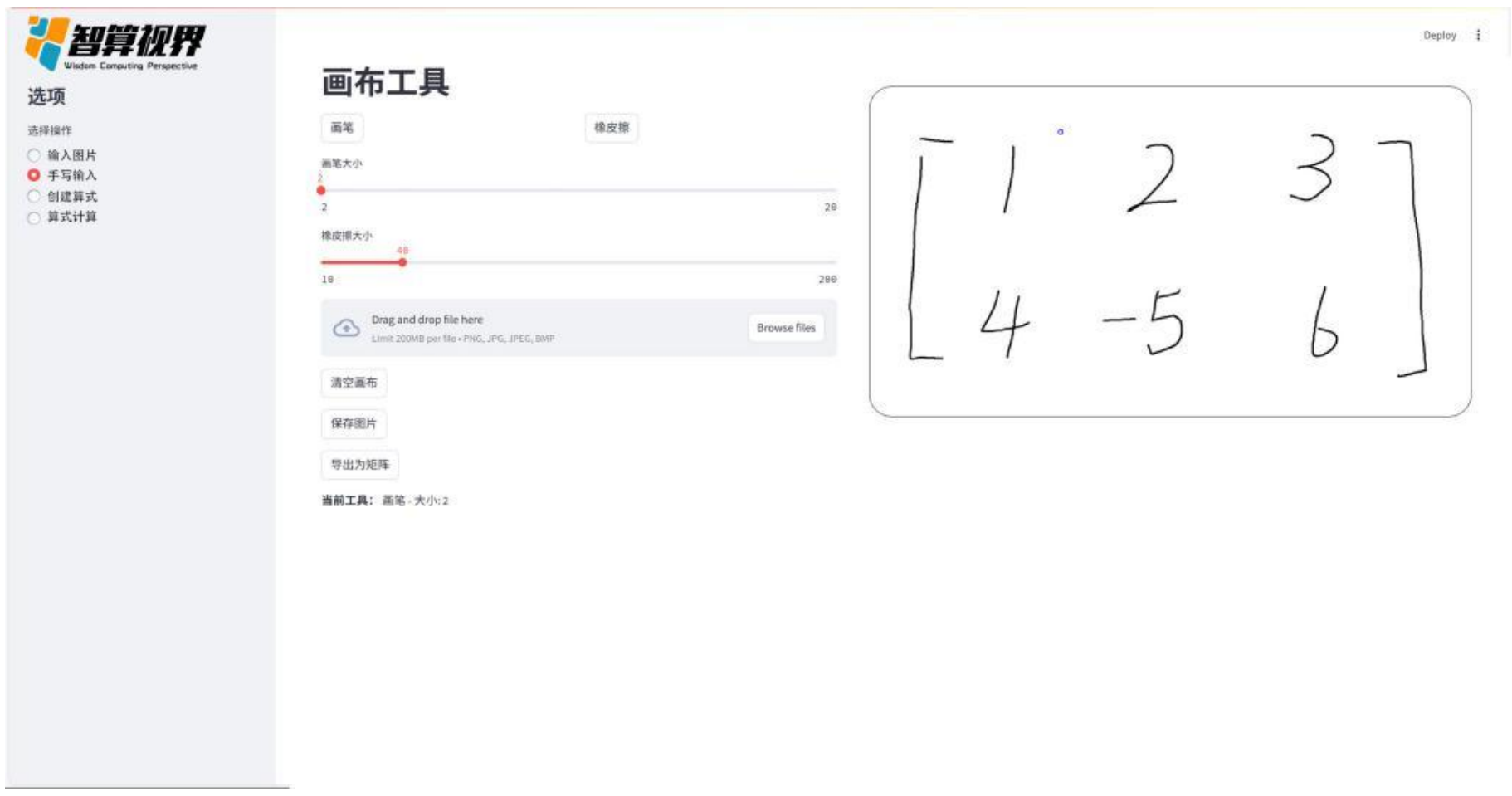

Figure 7.6, drawing board page

Innovatively added the undo and redo functions, implemented through two stacks. Add a list to record all paths drawn by a single click with the brush, then construct an operation stack to record the history of drawings, which is saved in the canvas_cache.txt file structure for easy direct reading pushpop and caching when opening next time. Build an undo stack to record all undone lists. When performing an undo operation, the undo stack pushes the current state onto the operation stack, and the entire canvas is redrawn based on Stack 1. When redoing, the opposite operations are performed. Save and export operations export the completed drawings from the entire stack, achieving the connection between the drawing board and the calculation matrix window. The specific algorithm flowchart is shown in Figure 7.7.

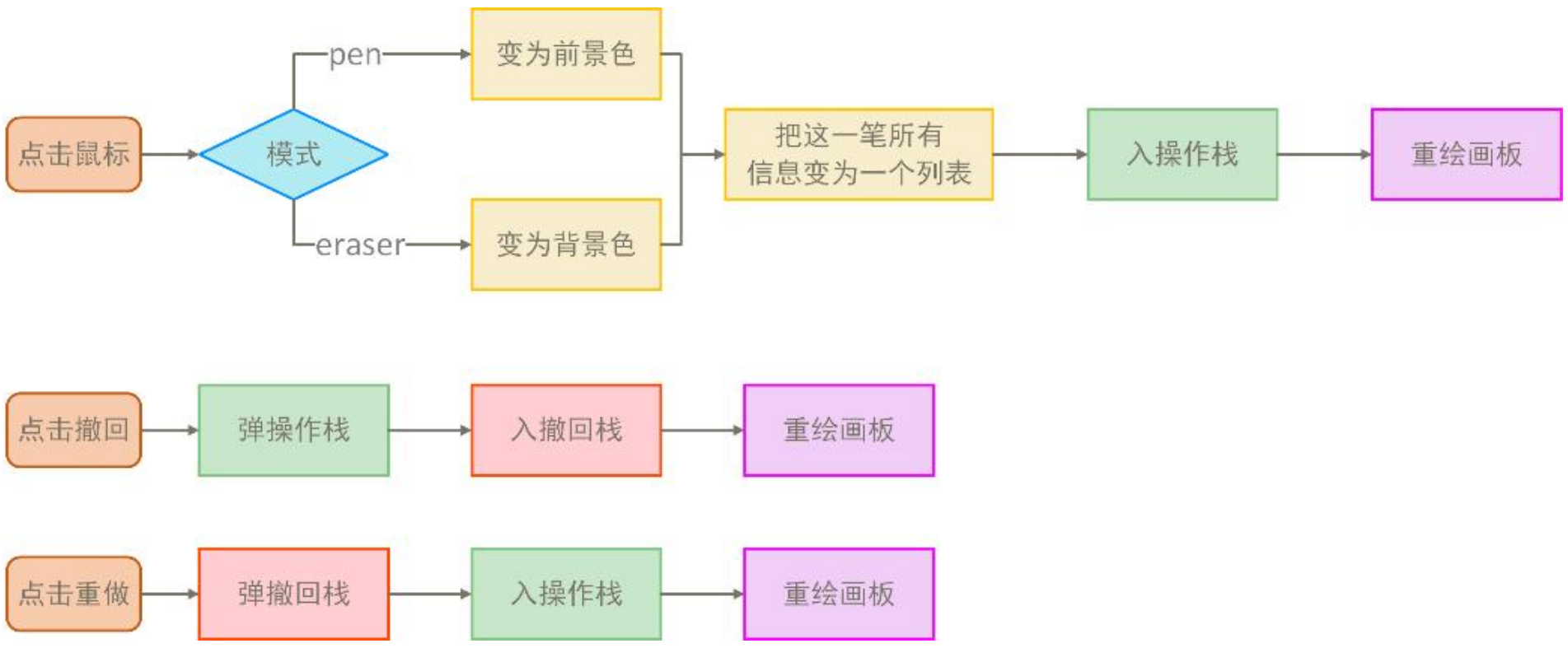

Figure 7.7 Flow chart of the drawing board algorithm

# 8 sum up

## 8.1sum up

This lesson is designed as an exploratory practice to deeply integrate artificial intelligence with education, providing more interactive and intuitive support for the teaching process. The project starts from common difficulties students face in understanding computational processes, attempting to replace "abstract formulas" with "visual processes." This approach breaks through the limitations of traditional teaching, which only presents calculation results without explaining intermediate steps. It ensures that each step of the calculation can be clearly displayed, overcoming the limitation of traditional calculation tools that focus solely on results while neglecting the expression of the process. This truly achieves the teaching goal of "making every step understandable to students."

During the development process, we adopted a lightweight YOLOv11 recognition model to achieve high-precision recognition of handwritten matrix images. We combined this with continuous training optimization using multiple version dataset generators, significantly enhancing the system's adaptability and robustness. In the animation module, we utilized the Manim animation engine to design and implement several custom classes, building a step-by-step animation logic for matrix addition, multiplication, determinants, and other operations from the ground up. This ensures that each mathematical calculation step has a clear and distinguishable visual presentation. Coupled with a graphical user interface and a user task caching mechanism, the system achieves a complete learning path loop from "handwriting input — recognition computation — animation output — homework submission," demonstrating strong engineering integration capabilities and educational adaptability.

As an educational tool that integrates handwriting recognition and dynamic visual demonstrations, it shows broad prospects in enhancing students 'comprehension, supporting teachers' teaching demonstrations, and promoting the digital transformation of textbooks. Three input methods (handwriting, photo-taking, editing) support different teaching scenarios. The step-by-step animation design aligns with the "process-oriented" teaching philosophy, offering flexible and user-friendly interaction. It can be used for classroom presentations, student self-practice, error feedback reconstruction, and more. This tool not only serves classroom explanations and post-class consolidation but can also integrate QR codes, micro-lecture videos, and other resource systems into textbooks, forming a tripartite teaching model of "print media + interaction + animation." Additionally, the system's micro-lecture generation capability provides significant potential support for future blended learning and online course resource development.

## 8.2look into the distance

The future development plan will mainly focus on the two core aspects of "deep expansion of mathematical calculation" and "intelligent upgrade of AI teaching aid", and strive to promote the dynamic visualization practice of artificial intelligence technology in teaching scenarios, which is embodied in the following six aspects:

1. Extended complex mathematical calculation support: On the basis of the original basic formula visualization, dynamic graphic demonstration of complex calculation processes such as matrix operation, linear algebra (such as eigenvalues, eigenvectors, matrix decomposition), calculus symbol derivation, function limit and derivative will be added in the future to help students understand abstract concepts more intuitively.

2. Construction of intelligent interactive teaching resource library: The system will continue to accumulate high-quality questions and teaching videos, and integrate interactive animation, formula derivation demonstration and instant practice feedback, so as to build a structured and hierarchical intelligent teaching resource library, providing accurate content matching for learners at different levels.

3. Development of teaching management and behavior visualization platform: For teachers, the teaching management background will be built to support real-time monitoring of students' learning behavior, visual trajectory analysis, error pattern recognition and progress curve tracking, so as to help teachers realize data-based accurate teaching and personalized guidance.

4. Introduction of AI teaching assistant and intelligent question answering system: Combined with natural language processing and knowledge graph, an AI teaching assistant with reasoning ability is developed, which can intelligently answer students' questions about math problems, give visual analysis process, and realize human-machine collaborative teaching.

5. Multi-platform and multi-scenario adaptive deployment: The system will realize cross-platform support, including PC, mobile terminal, touch device and interactive whiteboard, etc., to meet a variety of scenarios such as classroom teaching, after-school tutoring, independent learning, etc., and realize the intelligent learning experience of "learning anytime, anywhere".

6. Interdisciplinary integration and extended application: While consolidating the mathematical foundation, we also plan to gradually expand the system to physics, computer science and other disciplines highly related to mathematics, realize the dynamic visualization linkage of interdisciplinary knowledge, and provide intuitive and efficient AI solutions for more teaching scenarios.

# reference documentation